\documentclass{article}
\usepackage{newunicodechar}
\newunicodechar{≈}{\approx}

\usepackage[authoryear, sort&compress, round]{natbib}
\usepackage[preprint,logo]{agibot_enerverse}
\usepackage[T1]{fontenc}    
\usepackage{hyperref}       
\usepackage{url}            
\usepackage{amsfonts}       
\usepackage{nicefrac}       
\usepackage{microtype}      
\usepackage{amsmath}
\DeclareMathOperator*{\argmin}{argmin}
\usepackage{graphicx}
\usepackage{caption}
\usepackage{wrapfig}
\usepackage{pifont}
\usepackage[table]{xcolor}
\usepackage[absolute,overlay]{textpos}
\usepackage{algorithm}
\usepackage{algorithmic}
\usepackage[utf8]{inputenc}
\usepackage{booktabs}    
\usepackage{multirow}
\usepackage{makecell}
\usepackage{colortbl}
\usepackage{geometry}
\usepackage{siunitx}     
\usepackage{xcolor}
\definecolor{green}{RGB}{0,150,10}
\definecolor{blue}{RGB}{0,148,181}
\usepackage{threeparttable}
\definecolor{grey}{HTML}{999999}
\definecolor{lightblue}{HTML}{B0C4DE}
\definecolor{purple}{HTML}{E3BBED}
\definecolor{orange}{HTML}{ffdab9}
\definecolor{cadetblue}{HTML}{5F9EA0}
\definecolor{darksalmon}{rgb}{0.91, 0.59, 0.48}
\definecolor{forestgreen}{rgb}{0.13, 0.55, 0.13}
\definecolor{BlueGreen}{rgb}{0.0, 0.55, 0.55}
\definecolor{RedOrange}{rgb}{1.0, 0.27, 0.0}
\definecolor{sfiblue}{RGB}{230, 243, 255}
\definecolor{deltag}{RGB}{0, 140, 100}
\newcommand{\gain}[1]{\textbf{\textcolor{deltag}{{\scriptsize$\uparrow$}#1}}}
\definecolor{headerblue}{RGB}{41, 128, 185}
\definecolor{rowblue}{RGB}{235, 245, 251}
\definecolor{categoryrow}{RGB}{189, 215, 238}
\definecolor{ourwin}{RGB}{39, 174, 96}
\definecolor{summaryrow}{RGB}{255, 250, 205}
\definecolor{lightgray}{gray}{0.93}
\definecolor{speedupgreen}{HTML}{2E7D32}
\usepackage{xspace}

\title{Slow-Fast Inference: Training-Free Inference Acceleration via Within-Sentence Support Stability}

\author{%
  Xingyu Xie$^{1^*}$,\;
  Zhaochen Yu$^{1,2^*}$,\;
  Yue Liao$^{1^*}$,\;
  Tao Wang$^{2,\diamond}$,\;
  Kim-Chuan Toh$^{1,\diamond}$,\;
  Shuicheng Yan$^{1,\diamond}$\\[0.6em]
  \normalsize $^{1}$National University of Singapore \hspace{2mm} $^{2}$ByteDance\\[0.4em]
  \normalsize~\url{https://github.com/LV-NUS/SFI}
}

\begin{document}

\maketitle
\let\tempfootnote\thefootnote
\let\thefootnote\relax
\footnotetext{$^\diamond$ Corresponding Authors: Tao Wang (walton.wang929@gmail.com), Kim-Chuan Toh (mattohkc@nus.edu.sg) and Shuicheng Yan (yansc@nus.edu.sg).$^*$ Equal Contribution.}
\let\thefootnote\tempfootnote

\begin{abstract}
Long-context autoregressive decoding remains expensive because each decoding step must repeatedly process a growing history. We observe a consistent pattern during decoding: within a sentence, and more generally within a short semantically coherent span, the dominant attention support often remains largely stable. Motivated by this observation, we propose \textbf{Slow-Fast Inference} (SFI), a training-free decoding framework that decouples generation into frequent low-cost \textbf{fast steps} and occasional dense-attention \textbf{slow steps}. Fast steps reuse a compact sparse memory for efficient decoding. Slow steps are triggered near semantic boundaries. At slow steps, the model revisits the broader context and uses the \textbf{Selector} to refresh the selected memory for subsequent fast steps. Across the evaluated context lengths, SFI delivers approximately $1.6\times$--$14.4\times$ higher decoding throughput while generally maintaining quality on par with the full-KV baseline across long-context and long-CoT settings. Because SFI is training-free and applies directly to existing checkpoints, it offers a practical path to reducing inference cost for contemporary autoregressive reasoning models in long-context, long-horizon, and agentic workloads.
\end{abstract}

\section{Introduction}
\label{sec:intro}

Long-context inference has become a central workload for large language models (LLMs)~\citep{liu2025deepseek,yuan2025native, xie2025loco}.  This pressure is further amplified in modern long-sequence inference regimes. In retrieval-heavy applications, prefills can already span hundreds of thousands of tokens; in long chain-of-thought reasoning, the generated continuation may also become very long.  The trend is even more pronounced in emerging multi-agent systems, where the active context may accumulate prior agent messages, tool outputs, intermediate plans, and earlier reasoning traces~\citep{chen2026openclaw}.  As a result, both the retained context and the emitted reasoning trajectory can approach the practical context-window limit.

Although KV caching removes repeated key/value projections, each autoregressive decoding step still performs attention over the accessible history, incurring heavy compute and memory traffic as the context grows~\citep{chen2025magicpig, synk2025exploiting, xu2025refreshkv}.  In effect, the standard decoding pipeline still treats every step as a fresh reassessment of the entire past.  This tension motivates a basic question: \emph{does the model's attention focus truly reorganize at every token, or does it exhibit temporal structure that can be exploited for more efficient inference?}

\begin{figure}[t]
  \centering
\includegraphics[width=0.8\linewidth]{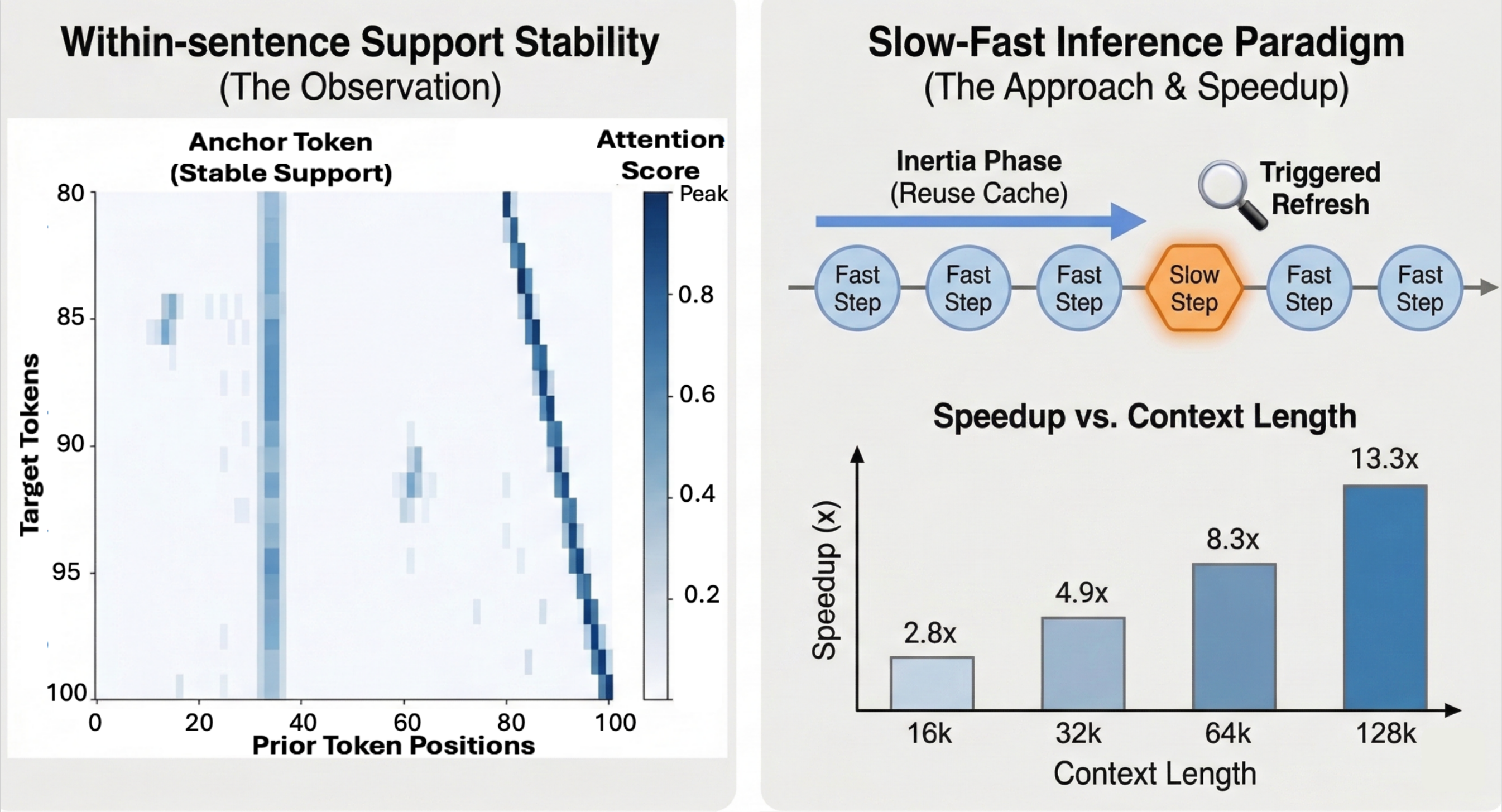}
  \caption{
  \textbf{The SFI Framework.}
\textbf{(A) Motivation:} Attention maps from Qwen3-0.6B illustrate a common pattern of \textbf{within-sentence support stability}: across consecutive decoding steps within a sentence, and more generally within a short semantically coherent span, the dominant attended positions remain largely stable rather than changing abruptly at every step. \textbf{(B) Slow-Fast paradigm and speedup:} SFI decouples decoding into many low-cost \textbf{Fast Steps} and occasional dense \textbf{Slow Steps}. The speedup plot reports the average end-to-end throughput gain across the Qwen series, from 0.6B to 235B, and shows that the advantage of this slow-fast schedule grows with context length.
}
  \label{fig:fig1}
  \vspace{-6mm}
\end{figure}

We observe that attention support often evolves more slowly than tokens are generated.  In particular, within a sentence, and more generally, within a short semantically coherent span, the model frequently assigns most attention mass to a largely overlapping subset of past positions, while larger support transitions are more likely to occur near semantic boundaries (Figure~\ref{fig:fig1}A;~\citealp{wu2025louiskv}).  We refer to this tendency as \emph{within-sentence support stability}. Although it does not hold uniformly at every step, it is frequent enough to motivate an event-driven decoding strategy: the model reuses a sparse support across multiple steps and refreshes it only when the attention support shows signs of shifting or when a preset reuse budget is exhausted.
Motivated by this structure, we propose Slow-Fast Inference (SFI), a training-free decoding framework that alternates frequent low-cost \textbf{fast steps} with occasional dense \textbf{slow steps} (Figure~\ref{fig:fig1}B). During fast steps, the model attends only to a compact cache composed of \textbf{sink + selected + recent} tokens: a small set of sink tokens that provide stable global anchors~\citep{xiao2024efficient}, a recent window updated at every step to preserve local continuity, and a selected memory that captures reusable long-range dependencies across multiple consecutive steps. During slow steps, the model revisits the accessible history with dense full attention and uses the resulting attention logits to refresh the selected memory. Slow steps are triggered primarily near sentence boundaries or other semantic transitions, and are additionally enforced once a fixed refresh budget is reached. This design concentrates dense full attention at positions where support transitions are more likely, while allowing most decoding steps to remain low-cost.

To turn dense full-attention evidence from each slow step into reusable sparse memory for subsequent fast steps, we introduce a training-free \textbf{Selector} that is invoked only during slow steps. Given the dense-attention logits collected at a slow step, the Selector combines (i) evidence from the current dense-attention observation and (ii) compact structural priors derived from cached statistics. We formulate this combination through a KL-based fusion objective with a closed-form solution, which produces a calibrated continuous score over the allowed positions. This score is then converted into sparse indices through score-space refinement followed by Top-$K$ discretization. In standard decoding, the evidence is taken from a single slow step (window size $=1$); during prefill, the same mechanism can optionally aggregate evidence over a short window for additional robustness.

This slow-fast decomposition yields speedups that grow with context length, because most decoding steps operate on a fixed-size sparse cache while dense refreshes occur relatively infrequently. To translate these algorithmic savings into end-to-end throughput gains, we further introduce two system-level designs: a latency-hiding asynchronous pipeline that overlaps refresh with decoding, and a memory-coalesced sparse-attention implementation that avoids bandwidth collapse under irregular sparse access.

Without any retraining and on the same model checkpoints, SFI attains near-parity quality with the full-KV baseline across long-context understanding and long-CoT settings, while improving decoding throughput by approximately $1.6\times$--$14.4\times$ in our efficiency evaluation. Because it requires no retraining and applies directly to existing checkpoints, SFI offers a practical path to reducing inference cost in long-context, long-horizon, and agentic workloads.
We summarize our main contributions as follows:
\begin{itemize}
  \item We identify \emph{within-sentence support stability}: during decoding, the dominant attention support often remains largely stable over short semantically coherent spans, and substantial support reconfiguration tends to occur near semantic boundaries.
  
  \item We propose SFI, a training-free decoding framework that decouples generation into frequent low-cost fast steps and occasional dense slow steps, reusing compact memory across multiple steps and refreshing it only when needed.
  
  \item We develop a training-free \textbf{Selector} that turns dense-attention evidence from slow steps into reusable selected memory through a KL-based fusion objective with a closed-form solution, followed by score-space refinement and discretization.
  
  \item We design an efficient system realization of SFI, including asynchronous slow-step overlap and a memory-coalesced sparse-attention kernel, and show that it preserves near-parity quality on long-context and long-CoT tasks while delivering approximately $1.6\times$--$14.4\times$ decoding throughput gains in practical inference settings.
\end{itemize}

\section{Related Work}
Work on efficient long-context inference can be broadly grouped into four directions: (i) inference-time cache selection and adaptive retention, which determine which past tokens remain directly accessible during decoding; (ii) budget allocation and related retention policies, which study where long-context capacity is most useful across layers or heads; (iii) orthogonal KV-cache compression, such as quantization and low-rank representations, which reduce memory footprint without changing the decoding schedule; and (iv) architecture and system co-design, which turns sparse access patterns into practical end-to-end gains. SFI is most closely related to the first direction, is complementary to the second, and relies on the fourth for practical acceleration. The third direction is largely orthogonal to our method and could in principle be combined with it. Accordingly, our empirical comparisons focus on representative training-free inference-time baselines that can be applied to the same existing checkpoints.

\textbf{Inference-Time Cache Selection and Adaptive Retention.}
A large body of work starts from the observation that attention during long-context decoding is often concentrated on a small subset of past tokens. Static methods compress the cache once after prefilling. SnapKV~\citep{li2024snapkv} selects clustered salient positions per head from an end-of-prompt observation window, while FastGen~\citep{ge2024model} identifies recurring attention patterns and applies corresponding retention rules. Dynamic methods instead update the accessible history during decoding. StreamingLLM~\citep{xiao2024efficient} retains attention sinks together with a recent-token window. H$_2$O~\citep{zhang2023h2o} keeps heavy-hitter tokens with high accumulated attention mass, while NACL~\citep{chen2024nacl} reduces the concentration bias of heavy-hitter retention through diversified random eviction. Scissorhands~\citep{liu2024scissorhands} further exploits the persistence of token importance across decoding steps. Another line avoids irreversible eviction by retrieving context on demand: Quest~\citep{tang2024quest} performs query-aware block retrieval, MagicPIG~\citep{chen2025magicpig} uses LSH-based importance sampling in a CPU--GPU heterogeneous design, and Loki~\citep{singhania2024loki} uses PCA-compressed keys for efficient top-$k$ retrieval. TidalDecode~\citep{yang2025tidaldecode} is also related in spirit: it exploits cross-layer coherence by using a few full-attention layers to identify salient positions that are then reused by later sparse layers.

SFI is closest to this family, but differs in its underlying decoding strategy. Existing methods typically improve efficiency by continuously evicting, updating, or retrieving context during decoding, or by reusing supports chosen by particular layers or approximate retrieval rules. SFI instead separates two questions: when full global context should be revisited, and how the reusable support for subsequent decoding steps should be constructed. At sparse slow steps, it recomputes exact full attention over the accessible history and uses the resulting dense evidence to build a head-wise token-level support for the following fast-step segment. This support is recallable rather than one-way: tokens omitted earlier can re-enter once the new dense evidence supports them. In this sense, SFI is neither pure eviction nor per-step retrieval, but an event-driven slow-fast decoding schedule built around exact dense refreshes and segment-level support reuse.

\textbf{Budget Allocation and Retention Policies.}
Another line of work studies not only which tokens to keep, but also where memory budget should be allocated across the network. At the layer level, PyramidKV~\citep{cai2024pyramidkv} reports a pyramidal information funneling effect, with broader attention in lower layers and more concentrated attention in upper layers, and allocates larger budgets to lower layers accordingly. DynamicKV~\citep{zhou2024dynamickv} further adapts layer-wise budgets on a per-input basis. At the head level, AdaKV~\citep{feng2024adakv} derives per-head allocation rules from an error bound on pruned attention outputs. DuoAttention~\citep{xiao2024duoattention} distinguishes retrieval heads from streaming heads, enabling full-span attention only where necessary, and HeadKV~\citep{fu2024headkv} jointly evaluates retrieval and reasoning roles to guide head-wise allocation.

These methods highlight that the utility of long-range memory can vary substantially across layers and heads. SFI is complementary to this perspective. Rather than redistributing budgets across network components or relying on learned specialization, SFI focuses on temporal reuse across decoding steps: it uses exact slow-step evidence to refresh a sparse support that can then be reused over multiple subsequent steps, while remaining fully training-free and model-agnostic.

\textbf{Orthogonal KV-Cache Compression.}
A separate direction compresses the KV cache itself. Quantization reduces the bit-width of stored keys and values, with the main challenge being activation outliers. KIVI~\citep{liu2024kivi} uses per-channel quantization for keys and per-token quantization for values. KVQuant~\citep{hooper2024kvquant} further proposes pre-RoPE key quantization, sparse outlier isolation, and full-precision retention for sink tokens. GEAR~\citep{kang2024gear} combines ultra-low-precision quantization with low-rank residual reconstruction and sparse outlier correction, while related techniques such as SmoothQuant~\citep{xiao2023smoothquant} and QuaRot~\citep{ashkboos2024quarot} aim to make aggressive compression more stable. Low-rank methods exploit redundancy in KV representations. LoRC~\citep{zhang2024lorc}, Palu~\citep{chang2024palu}, and ShadowKV~\citep{sun2024shadowkv} apply low-rank decomposition to KV-related components, while MLA~\citep{deepseek2024} compresses keys and values into latent representations at the architectural level.
These approaches mainly reduce the memory footprint of each retained token, whereas \emph{SFI reduces how often expensive dense full-history attention must be executed}. The two directions are therefore largely orthogonal and, in principle, can be combined.

\textbf{Hybrid Sparse Attention Architectures.}
Recent models such as Gemma~3~\citep{gemma2025} and MiMo-V2-Flash~\citep{xiaomi2026mimo} interleave sliding-window and global-attention layers, reducing KV-cache growth by restricting full-context access to only part of the network. HySparse~\citep{gao2026hysparse} pushes this idea further by using full-attention layers as token-selection oracles whose outputs are reused by subsequent sparse layers. These designs are related in spirit to SFI in that they also separate cheaper local processing from more expensive global access. However, they typically depend on model-specific sparse/global layer patterns, oracle layers, or other architecture-level design choices, and in some cases also involve training or adaptation. By contrast, SFI is a post hoc inference-time method that applies to existing checkpoints without introducing new layer patterns or structural changes. We therefore discuss these approaches for context rather than treat them as primary head-to-head baselines.

\textbf{System and Kernel Co-Design.}
Even when sparse attention reduces nominal computation, practical speedups are not automatic. Sparse decoding often suffers from irregular memory gathers, poor coalescing, runtime synchronization overhead, and control-flow fragmentation. More broadly, recent system studies~\citep{zhang2025sageattention,zhang2024sageattention2,xie2024adan,xie2024optimization} show that the gap between algorithmic sparsity and wall-clock acceleration is often determined by memory layout, kernel design, and scheduling rather than FLOPs alone. Our infrastructure design follows the same principle, but is specialized to the slow-fast execution pattern: the sparse fast path must remain bandwidth-efficient, while dense refresh and cache maintenance must be overlapped and amortized carefully to translate the algorithmic idea into robust end-to-end throughput gains.

\section{Method}
\label{sec:method}
We now formalize SFI. The core idea is to alternate low-cost fast steps, which attend to a managed sparse state, with occasional slow steps, which run dense full attention to refresh the selected memory for the next segment. Both the trigger policy and the cache update rule are determined online without retraining: a slow step exposes masked attention logits over an allowed candidate set, and the \textbf{Selector} maps these logits, together with lightweight cache statistics, to Top-$K$ index updates via a KL-based fusion objective with an exact closed-form solution.

In the remainder of this section, we first formalize the managed sparse state and the trigger policy that governs the transition between fast and slow steps (Sec.~\ref{sec:framework}).
We then present the Selector that updates long-range memory from slow-step logits via closed-form fusion (Sec.~\ref{sec:selector}),
followed by the score-space refinements, Top-$K$ discretization, and complexity discussion (Sec.~\ref{sec:refine_topk}).

\begin{figure*}[t]
  \begin{center}
    \centerline{\includegraphics[width=0.9\linewidth]{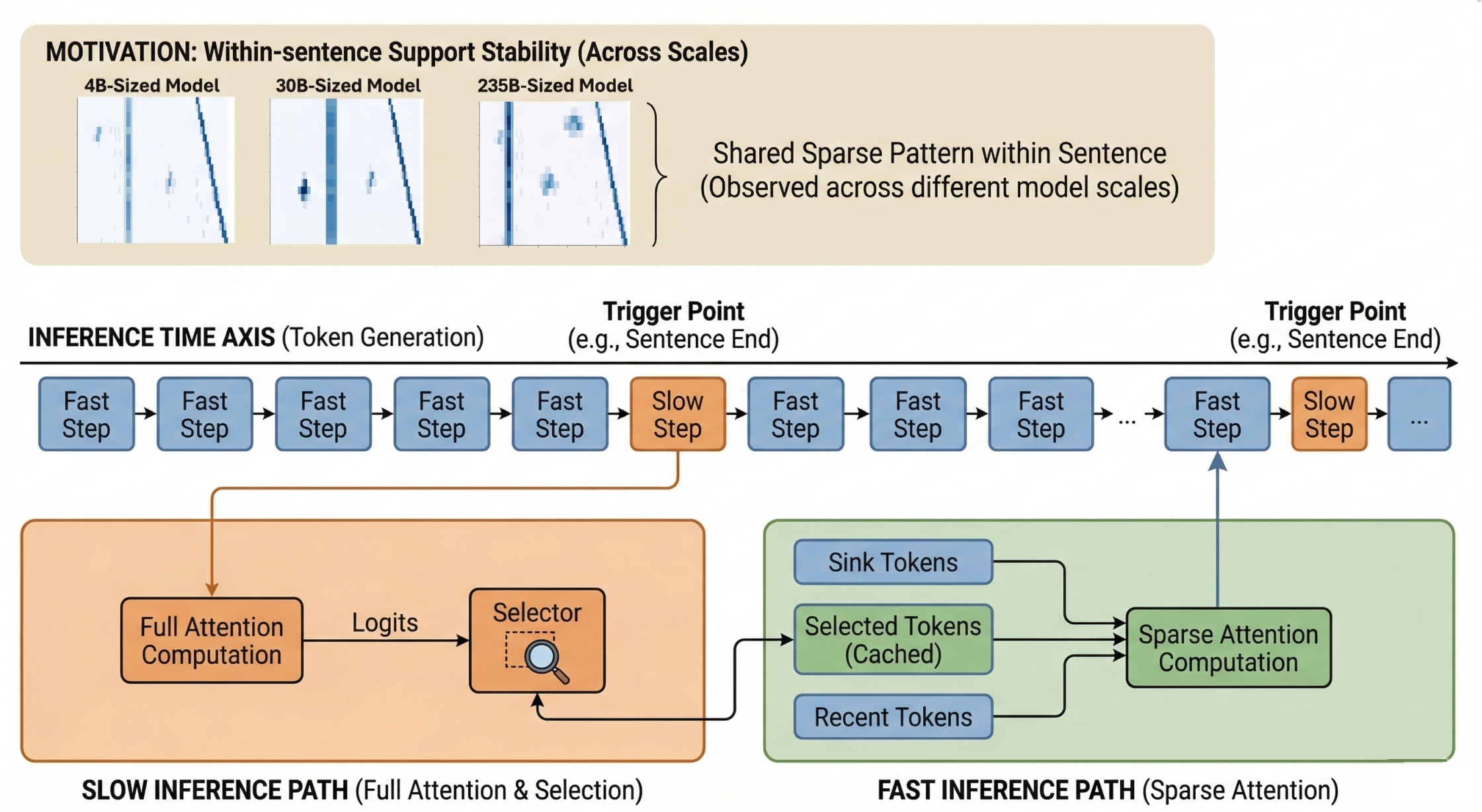}}
    \caption{
    \textbf{The Slow-Fast Inference framework.}
    \textbf{Top:} Across multiple model scales, attention mass often remains concentrated on a largely stable set of positions within a semantic unit, illustrating \textbf{within-sentence support stability}.
    \textbf{Bottom:} SFI exploits this pattern by alternating frequent low-cost \textbf{fast steps}, which attend to a managed sparse state (\textit{sink + selected + recent}), with occasional \textbf{slow steps}. A slow step is triggered when a boundary token is generated (Eq.~\eqref{eq:trigger}) or when a fixed refresh interval is reached; it then performs dense attention, collects masked attention logits over the allowed candidate set, and invokes the \textbf{Selector} to refresh the selected memory for the next segment.
    }
    \label{fig:fig2}
  \end{center}
\end{figure*}

\subsection{Slow-Fast Inference Framework}
\label{sec:framework}

We now describe the decoding state maintained by SFI, how it is reused in fast steps and refreshed in slow steps, and the training-free trigger policy that schedules these refreshes.

\paragraph{Managed sparse state.}
Let $L_t$ be the prefix length at generation step $t$.
For each layer, we maintain a sparse index set $\mathcal{I}^{(t)}$ that defines the KV entries accessed by fast steps:
\begin{equation}
  \mathcal{I}^{(t)}
  =
  \underbrace{\mathcal{I}_{\mathrm{sink}}}_{\text{anchors}}
  \ \cup\
  \underbrace{\mathcal{I}_{\mathrm{recent}}^{(t)}}_{\text{local}}
  \ \cup\
  \underbrace{\mathcal{I}_{\mathrm{sel}}^{(t)}}_{\text{selected}}.
  \label{eq:index_set}
\end{equation}
This state has three roles. $\mathcal{I}_{\mathrm{sink}}$ is a small fixed set of anchor tokens (e.g., prefix/sink tokens) that stabilizes attention; $\mathcal{I}_{\mathrm{recent}}^{(t)}$ is a sliding window that preserves short-range dependencies; and $\mathcal{I}_{\mathrm{sel}}^{(t)}$ stores the selected memory, namely the reusable long-range tokens carried across consecutive fast steps, and is the only subset updated by the Selector. For notational clarity, we present a single set per layer; in practice it is maintained per KV head, and in GQA~\citep{ainslie2023gqa} it is natural to share the set across heads that share keys and values. Accordingly, the fast-step attention cost scales with $|\mathcal{I}^{(t)}|$ rather than $L_t$.

To refresh the long-range component, we define an allowed candidate set that excludes the mandatory sink and recent entries:
\begin{equation}
  \mathcal{J}^{(t)} = \{1,\dots,L_t\}\setminus\!\Big(\mathcal{I}_{\mathrm{sink}}\cup \mathcal{I}_{\mathrm{recent}}^{(t)}\Big),
  \label{eq:allowed_set}
\end{equation}
so the Selector only ranks positions in $\mathcal{J}^{(t)}$ and never competes with sink/recent tokens. Equivalently, $\mathcal{J}^{(t)}$ can be viewed as a binary mask applied to the logits produced by a slow step.

\paragraph{Fast step.}
Given the managed sparse state above, a fast step at time $t$ computes attention only over the keys and values indexed by $\mathcal{I}^{(t)}$. During this phase, the selected long-range memory is reused rather than recomputed:
\begin{equation}
  \mathcal{I}_{\mathrm{sel}}^{(t+1)}=\mathcal{I}_{\mathrm{sel}}^{(t)}
  \qquad \text{(Fast step)}.
  \label{eq:freeze_sel}
\end{equation}
Meanwhile, $\mathcal{I}_{\mathrm{recent}}^{(t)}$ slides with $t$ to track the immediate local context. This is the low-cost path that SFI executes for most decoding steps.

\paragraph{Slow step.}
The role of a slow step is to refresh the long-range part of the sparse state when the current support is no longer reliable. Near sentence boundaries or segment transitions, globally relevant context can shift and invalidate the previously selected memory. A slow step therefore performs dense attention over the prefix and records a short observation window of masked attention logits over $\mathcal{J}^{(t)}$. Specifically, we collect logits for a window of $W$ queries (e.g., the last $W$ queries in a prefill block; in decode we typically use $W=1$ for efficiency):
\begin{equation}
  \big\{\ell_{\tau}(j)\big\}_{\tau=1}^{W,\ j\in \mathcal{J}^{(t)}},
  \label{eq:window_logits}
\end{equation}
where $\ell_{\tau}(j)$ denotes the masked attention logit assigned to position $j$ by the $\tau$-th query in the window. These logits, together with lightweight cached statistics such as key norms and positions, are passed to the \textbf{Selector}, which returns an updated selected set $\mathcal{I}_{\mathrm{sel}}^{(t+1)}$ implementing the refreshed selected memory. Subsequent fast steps then reuse this refreshed memory until the next slow step.

\paragraph{Training-free trigger policy.}
The remaining question is when to switch from the cheap reuse path to a slow step. We use a simple boundary-based rule specified by a user-defined set of trigger token IDs $\mathcal{T}_{\mathrm{trig}}$ (e.g., sentence-ending punctuation, paragraph separators, or heading markers).
Let $x_t$ denote the token generated at decoding step $t$, and let
$g_t\in\{0,1\}$ indicate whether the computation that generates $x_t$
is a slow step. We decide the type of the current step from the token
generated at the previous step: if the last token is a trigger token,
the next decoding step becomes slow. Formally, for $t\ge 1$,
\begin{equation}
  g_t = \mathbb{I}\!\left[x_{t-1} \in \mathcal{T}_{\mathrm{trig}}\right],
  \label{eq:trigger}
\end{equation}
where $\mathbb{I}[\cdot]$ is the indicator function. The set
$\mathcal{T}_{\mathrm{trig}}$ is constructed offline via the tokenizer,
so runtime detection operates directly on token IDs without string
decoding. To prevent the selected memory from remaining stale for too
long, we additionally force $g_t=1$ if no slow step has occurred for
$T_{\max}$ decoding steps. 
In our default setting, $\mathcal{T}_{\mathrm{trig}}$ targets sentence or segment boundaries, where support shifts are most likely.

\subsection{Closed-form Selector via Reverse-KL Fusion}
\label{sec:selector}

At each slow step (Sec.~\ref{sec:framework}), we run dense attention and record masked attention logits for a short query window of length $W$.
The \textbf{Selector} converts this dense observation into a per-head continuous importance distribution over allowed key positions; Sec.~\ref{sec:refine_topk} then discretizes it into Top-$K$ indices to update the sparse KV state.

We consider one layer and one KV head (head index omitted when clear).
Let $\mathcal{J}$ denote the allowed set defined in Sec.~\ref{sec:framework}, and let
\[
  \Delta_{\mathcal{J}}
  =\Big\{ s:\mathcal{J}\to\mathbb{R}_{\ge 0}\ \Big|\ \sum_{j\in\mathcal{J}} s(j)=1 \Big\}
\]
be the probability simplex on $\mathcal{J}$.
Throughout, $\varepsilon>0$ is a fixed small constant for numerical stability (e.g., $\varepsilon=10^{-8}$), used to avoid undefined operations such as division by zero and powers at zero.

\paragraph{Evidence from slow-step logits.}
For each query step $t\in\{1,\dots,W\}$, we convert masked logits $\{\ell_t(j)\}_{j\in\mathcal{J}}$ into an attention distribution on $\mathcal{J}$:
\begin{equation}
  p_t(j)
  =\operatorname{Softmax}\!\big(\{\ell_t(i)\}_{i\in\mathcal{J}}\big)(j)
  =\frac{\exp(\ell_t(j))}{\sum_{i\in\mathcal{J}}\exp(\ell_t(i))},
  \qquad j\in\mathcal{J},
  \label{eq:selector_pt}
\end{equation}
with $p_t(j)=0$ for $j\notin\mathcal{J}$ under the mask.
We summarize the window into a nonnegative evidence score by a power-mean statistic ($\alpha\in(0,1]$):
\begin{equation}
  \mu(j)
  =\frac{1}{W}\sum_{t=1}^{W} p_t(j)^{\alpha},
  \qquad j\in\mathcal{J}.
  \label{eq:selector_mu}
\end{equation}
Choosing $\alpha<1$ applies a power-transform smoothing in log-probability space, so the window summary better reflects the support that receives nontrivial mass rather than being dominated by a few very large entries.
When $W=1$ (the common decode case), Eq.~\eqref{eq:selector_f} below recovers $f=p_1$ exactly.

The vector $\mu$ is not normalized, so we convert it into a probability distribution on $\mathcal{J}$ for subsequent probabilistic fusion.
Since the aggregation is performed in the $\alpha$-power domain, we first apply the inverse map and then normalize (an $\ell_1$ projection onto $\Delta_{\mathcal{J}}$):
\begin{equation}
  f(j)
  =\frac{\mu(j)^{1/\alpha}}{\sum_{i\in\mathcal{J}}\mu(i)^{1/\alpha}},
  \qquad j\in\mathcal{J}.
  \label{eq:selector_f}
\end{equation}
This mapping preserves the relative evidence implied by $\mu$ (it is monotone in $\mu$), removes arbitrary scale variation in $\mu$, and yields a valid distribution on the same support as the prior.

\paragraph{Cache-aware prior on allowed positions.}
Because the slow-step observation window is short, the evidence $f$ can become overly concentrated on a small set of positions, which may hurt long-range coverage. We therefore introduce a lightweight prior $r\in\Delta_{\mathcal{J}}$, computed directly from cached statistics without additional forward passes. The prior is designed to counter two biases that are particularly pronounced under short-window estimation: logit inflation caused by unusually large key norms, and over-concentration on very recent positions near the tail of the allowed range.

\textbf{Key-norm factor.}
Let $\|k(j)\|_2$ denote the cached key norm at position $j$. Since attention logits are formed from query--key inner products, a key with an unusually large norm can receive a high logit even when its directional match with the current query is not especially strong~\citep{devoto2024simple}. When the evidence is estimated from only a short window, this effect can make a few large-norm positions appear spuriously important and crowd out other useful long-range candidates. To reduce this bias, we introduce a heavy-tailed downweighting factor
\begin{equation}
  \pi_{\mathrm{kn}}(j)\propto (\|k(j)\|_2+\varepsilon)^{-\gamma},
  \qquad \gamma\ge 0,
  \label{eq:selector_prior_kn}
\end{equation}
which softly penalizes unusually large-norm keys while still preserving non-negligible mass on them when the evidence truly supports their selection. In this way, the prior suppresses norm-driven outliers without eliminating genuinely informative positions.

\textbf{Position factor (global decay + tail brake).}
Short-window evidence can also become overly concentrated on the most recent part of the allowed range, especially when a slow step is triggered near a semantic transition. If left uncorrected, Top-$K$ may repeatedly spend budget on a narrow near-tail region, leading to redundant selections and poorer temporal coverage over the prefix. To alleviate this bias, we introduce a smooth position prior with two roles: a global decay that prevents the score mass from collapsing toward recent positions, and an additional tail brake that specifically suppresses over-concentration at the extreme end.

Let $u(j)\in[0,1]$ denote the normalized position of $j$ within $\mathcal{J}$, where larger $u(j)$ corresponds to more recent positions among the allowed keys:
\begin{equation}
  u(j)=\frac{j-j_{\min}}{j_{\max}-j_{\min}+\varepsilon},
  \qquad
  j_{\min}=\min\mathcal{J},\ \ j_{\max}=\max\mathcal{J}.
  \label{eq:selector_u}
\end{equation}
We then define
\begin{equation}
  \pi_{\mathrm{pos}}(j)\propto
  \exp\!\big(-\beta\,u(j)^p\big)\cdot (1-u(j)+\varepsilon)^{\eta},
  \qquad \beta\ge 0,\ p\ge 1,\ \eta\ge 0.
  \label{eq:selector_prior_pos}
\end{equation}
The first factor, $\exp(-\beta u(j)^p)$, provides a smooth global discount on recency, encouraging the prior to distribute probability mass over a broader temporal range rather than concentrating too heavily near the tail. Here, $\beta$ controls the overall strength of the decay, while $p$ controls its curvature. The second factor, $(1-u(j)+\varepsilon)^{\eta}$, is relatively mild over most of the range but drops more sharply as $u(j)\to 1$, so it mainly acts on the extreme tail where redundancy is most likely. Importantly, this prior does not rule out recent tokens; it only reduces their tendency to dominate the selected set when the short-window evidence is overly tail-concentrated.
We combine the key-norm and position factors and then normalize:
\begin{equation}
  r(j)
  =
  \frac{\pi_{\mathrm{kn}}(j)\,\pi_{\mathrm{pos}}(j)}
  {\sum_{i\in\mathcal{J}}\pi_{\mathrm{kn}}(i)\,\pi_{\mathrm{pos}}(i)},
  \qquad j\in\mathcal{J}.
  \label{eq:selector_r}
\end{equation}
\paragraph{Reverse-KL fusion of evidence and prior.}
We now have two distributions defined on the same support $\mathcal{J}$: the evidence distribution $f$, derived from dense attention in the slow-step window, and the prior distribution $r$, which encodes cache-aware regularities. Neither is sufficient on its own. Using $f$ alone can make the selection overly narrow and sensitive to short-window noise, while using $r$ alone would ignore the current slow-step observation. We therefore seek a fused distribution $s_{\lambda}\in\Delta_{\mathcal{J}}$ that remains faithful to the current evidence while being regularized by the prior. This fused distribution serves as the calibrated continuous score from which the final sparse support will be selected.

We define $s_{\lambda}$ by minimizing a convex combination of Kullback-Leibler (KL) divergences:
\begin{equation}
  s_{\lambda}
  =
  \argmin_{s\in\Delta_{\mathcal{J}}}
  (1-\lambda)\,D_{\mathrm{KL}}(f\|s)
  +\lambda\,D_{\mathrm{KL}}(r\|s),
  \qquad \lambda\in[0,1],
  \label{eq:selector_obj}
\end{equation}
where $D_{\mathrm{KL}}(a\|b)=\sum_{j\in\mathcal{J}} a(j)\log\!\frac{a(j)}{b(j)}$ for $a,b\in\Delta_{\mathcal{J}}$. The parameter $\lambda$ controls the trade-off: smaller $\lambda$ keeps $s_{\lambda}$ closer to the slow-step evidence $f$, while larger $\lambda$ places more weight on the prior $r$.
This objective also admits an exact closed-form solution. Expanding each KL term gives
\[
  D_{\mathrm{KL}}(a\|s)
  =\sum_{j\in\mathcal{J}} a(j)\log a(j)-\sum_{j\in\mathcal{J}} a(j)\log s(j),
\]
where the first term is constant with respect to $s$. Therefore, Eq.~\eqref{eq:selector_obj} is equivalent to
\begin{equation}
  \max_{s\in\Delta_{\mathcal{J}}}\;
  \sum_{j\in\mathcal{J}} w_{\lambda}(j)\log s(j),
  \qquad
  w_{\lambda}(j)=(1-\lambda)f(j)+\lambda r(j).
  \label{eq:selector_obj_equiv}
\end{equation}
Introducing a Lagrange multiplier for the simplex constraint $\sum_j s(j)=1$, the stationary condition gives $s(j)\propto w_{\lambda}(j)$. Since $w_{\lambda}$ is already normalized, the solution is immediate:
\begin{equation}
  s_{\lambda}(j)=(1-\lambda)\,f(j)+\lambda\,r(j),
  \qquad j\in\mathcal{J}.
  \label{eq:selector_closed}
\end{equation}
Thus, the fused score is simply a convex interpolation between current slow-step evidence and the cache-aware prior, yielding a continuous ranking signal that is both adaptive and regularized before the final Top-$K$ discretization.

\paragraph{Why reverse KL instead of forward KL.}
Eq.~\eqref{eq:selector_obj} (reverse KL) yields an arithmetic mixture.
For comparison, the forward-KL alternative
\[
  \min_{s\in\Delta_{\mathcal{J}}}(1-\lambda)D_{\mathrm{KL}}(s\|f)+\lambda D_{\mathrm{KL}}(s\|r)
\]
yields a geometric mixture $s(j)\propto f(j)^{1-\lambda}r(j)^{\lambda}$.
The distinction becomes most evident when we consider a single position $j$.
If $f(j)$ assigns substantial mass but $r(j)$ is small, then the geometric term $f(j)^{1-\lambda}r(j)^{\lambda}$ becomes small because it is multiplicative in the two sources; in other words, geometric fusion keeps a position large only when both sources agree.
In contrast, the arithmetic mixture in Eq.~\eqref{eq:selector_closed} retains a contribution from either source through addition, which is more compatible with candidate selection where discarding a useful token (false negative) can be more damaging than retaining a small number of extra candidates.
A pointwise inequality summarizes this relation:
\begin{equation}
  (1-\lambda)f(j)+\lambda r(j)\;\ge\; f(j)^{1-\lambda}r(j)^{\lambda},
  \qquad \forall j\in\mathcal{J}.
  \label{eq:selector_amgm}
\end{equation}

\paragraph{Choosing $\lambda$ by discretization stability.}
Eq.~\eqref{eq:selector_closed} gives a one-parameter family of fused scores that trades off current evidence and prior. The subsequent discretization depends only on the ranking induced by $s_{\lambda}$, so a useful choice of $\lambda$ should make this ranking less sensitive to small estimation errors in $f$. In particular, when the fused score is too sharp, small perturbations in the evidence can cause large changes among the borderline entries of the selected set. We therefore choose $\lambda$ to make $s_{\lambda}$ more stable, while limiting the amount of prior injection through $\lambda_{\text{clip}}$:
\begin{equation}
  \lambda^*
  =\argmin_{\lambda\in[0,\lambda_{\text{clip}}]}\ \|s_{\lambda}\|_2^2,
  \qquad s=s_{\lambda^*},
  \label{eq:selector_lambda_opt}
\end{equation}
where $\|s\|_2^2=\sum_{j\in\mathcal{J}} s(j)^2$ penalizes sharp distributions. Minimizing this quantity encourages the fused score to spread mass across multiple plausible positions instead of placing excessive weight on only a few entries, which in turn makes the induced ranking more robust.
The minimizer admits the closed form
\begin{equation}
  \lambda^*
  =
  \mathrm{clip}\!\left(
    \frac{\|f\|_2^2-f^\top r}{\|f\|_2^2-2f^\top r+\|r\|_2^2},
    \ 0,\ \lambda_{\text{clip}}
  \right),
  \qquad
  s=(1-\lambda^*)f+\lambda^* r,
  \label{eq:selector_lambda_closed}
\end{equation}
where $\mathrm{clip}(x,0,\lambda_{\text{clip}})=\min\{\max\{x,0\},\lambda_{\text{clip}}\}$. The resulting $s$ is the final continuous importance score on $\mathcal{J}$, which is converted into the updated selected set in the next stage.

\subsection{Log-score Refinements and Top-$K$ Discretization}
\label{sec:refine_topk}

Sec.~\ref{sec:selector} produces, for each head, a continuous importance distribution $s_h\in\Delta_{\mathcal{J}}$ on the allowed set $\mathcal{J}$. The goal of this stage is to convert $s_h$ into Top-$K$ indices for updating the selected memory. A direct Top-$K$ on $s_h$ is often suboptimal for two reasons: within a head, several nearby positions may all receive high scores because of local score correlation; across heads, the same position may be selected repeatedly. To reduce these two forms of redundancy, we apply two lightweight refinements in \textbf{log-score space} before discretization. Since Top-$K$ depends only on score ordering, these refinements do not need to preserve normalization or define a probability distribution. Concretely, we start from
\begin{equation}
  z_h(j)=\log\!\big(s_h(j)+\varepsilon\big),\qquad j\in\mathcal{J},
  \label{eq:z_base}
\end{equation}
and use $[x]_+=\max(x,0)$ below.

\textbf{Soft non-maximum suppression (Soft-NMS).}
The first refinement operates within each head. The issue is that when one local region contains several highly correlated candidates, naive Top-$K$ may allocate multiple slots to that same neighborhood, even though these positions carry very similar information. Our goal is to preserve the local maximum in each neighborhood while decaying nearby lower-scoring candidates. In this way, one narrow region is less likely to occupy multiple Top-$K$ slots, and high-quality candidates from other neighborhoods have a better chance to enter the selected set, improving positional coverage.

Let $\mathcal{N}(j)$ denote a small neighborhood around position $j$. We first compute the local maximum
\begin{equation}
  m_h(j)=\max_{i\in\mathcal{N}(j)} z_h(i),
  \label{eq:nms_pool}
\end{equation}
and then apply a gap-based decay
\begin{equation}
  z_h^{\mathrm{nms}}(j)
  =z_h(j)-\alpha_{\text{soft}}\,[m_h(j)-z_h(j)]_+,
  \qquad \alpha_{\text{soft}}\ge 0.
  \label{eq:nms}
\end{equation}
This update has a simple interpretation. If $j$ is a local maximizer in its neighborhood, then $m_h(j)=z_h(j)$ and its score is unchanged. If $j$ is weaker than a nearby competitor, then the gap $m_h(j)-z_h(j)$ is positive, and its score is reduced proportionally. As a result, the dominant candidate in each neighborhood remains competitive, while nearby non-maximal positions are pushed down. This discourages Top-$K$ from spending many slots on one narrow region and allows strong candidates from other neighborhoods to remain in contention.

\textbf{Cross-head exclusivity.}
The second refinement operates across heads. Even after local suppression, different heads may still rank the same position highly, which reduces diversity in the selected memory. To mitigate this, we introduce a soft competition across heads at each fixed position.

For each position $j\in\mathcal{J}$, define the head-wise score vector
\[
  \mathbf{z}^{\mathrm{nms}}(j)
  =
  \big[z_{1}^{\mathrm{nms}}(j),\dots,z_{H}^{\mathrm{nms}}(j)\big],
\]
where $H$ is the number of KV heads. We convert these scores into a head-wise responsibility assignment using
\begin{equation}
  r_h(j)
  =
  \operatorname{Softmax}\!\big(\mathbf{z}^{\mathrm{nms}}(j)/T\big)_h,
  \qquad T>0,
  \label{eq:head_softmax}
\end{equation}
where a smaller temperature $T$ yields sharper competition and a larger $T$ makes the assignment softer. We then adjust the log-scores by
\begin{equation}
  z_h^{\mathrm{adj}}(j)
  =
  z_h^{\mathrm{nms}}(j)
  +\alpha_{\text{cross}}\,\log\!\big(\max(r_h(j),\varepsilon)\big),
  \qquad \alpha_{\text{cross}}\ge 0.
  \label{eq:head_mutex}
\end{equation}
Because $\log r_h(j)< 0$, this is always a negative adjustment. A head that is strongly responsible for position $j$ has $r_h(j)\approx 1$, so its score changes little. By contrast, heads with small responsibility at the same position receive a larger penalty. In effect, once one head strongly claims a position, competing heads are softly discouraged from selecting it as well, making it more likely that their Top-$K$ selections move to alternative positions. The parameter $\alpha_{\text{cross}}$ controls the strength of this exclusivity effect.

\paragraph{Top-$K$ output and state update.}
We discretize using the refined scores and update the selected indices:
\begin{equation}
  S_h=\mathrm{TopK}_{j\in\mathcal{J}}\big(z_h^{\mathrm{adj}}(j),K\big),
  \qquad
  \mathcal{I}_{\mathrm{sel},h}^{(t+1)} \leftarrow S_h,
  \label{eq:topk_out}
\end{equation}
and form the sparse attention index set for subsequent fast steps as
\begin{equation}
  \mathcal{I}_h^{(t+1)}
  =
  \mathcal{I}_{\mathrm{sink}}
  \cup
  \mathcal{I}_{\mathrm{recent}}^{(t+1)}
  \cup
  \mathcal{I}_{\mathrm{sel},h}^{(t+1)}.
  \label{eq:index_set_head}
\end{equation}

\paragraph{Complexity.}
Let $H$ be the number of KV heads.
Soft-NMS and cross-head exclusivity each apply simple reductions and pointwise transforms over $\mathcal{J}$, requiring $O(H|\mathcal{J}|)$ work per layer.
In practice, the dominant cost within discretization is typically the per-head Top-$K$ over $|\mathcal{J}|$ scores; the refinement operators add only a small overhead compared to dense attention in a slow step.

\begin{algorithm}[t]
  \caption{Slow-Fast Inference (SFI) loop for one request}
  \label{alg:sfi_loop}
  \begin{algorithmic}[1]
    \REQUIRE Anchor indices $\mathcal{I}_{\mathrm{sink}}$; sliding-window rule for $\mathcal{I}_{\mathrm{recent}}^{(t)}$; trigger set $\mathcal{T}_{\mathrm{trig}}$; refresh budget $T_{\max}$; selection budget $K$; window length $W$
    \ENSURE Generated tokens and sparse states induced by Eq.~\eqref{eq:index_set_head}

    \STATE Initialize $\mathcal{I}_{\mathrm{recent}}^{(0)}$ from the prompt context and $\mathcal{I}_{\mathrm{sel},h}^{(0)}$ for all heads $h$
    \STATE Set $t \leftarrow 0$ and $g_0 \leftarrow 1$

    \WHILE{generation is not finished}
      \IF{$g_t = 0$}
        \STATE \textbf{Fast Step:} attend with
        $
          \mathcal{I}_{\mathrm{sink}}
          \cup
          \mathcal{I}_{\mathrm{recent}}^{(t)}
          \cup
          \mathcal{I}_{\mathrm{sel},h}^{(t)}
        $
        for each head $h$
        \STATE $\mathcal{I}_{\mathrm{sel},h}^{(t+1)} \leftarrow \mathcal{I}_{\mathrm{sel},h}^{(t)}$ for all heads $h$
      \ELSE
        \STATE \textbf{Slow Step:} run dense attention and record window logits $\{\ell_{\tau}(j)\}_{\tau=1..W,\ j\in\mathcal{J}^{(t)}}$ as in Eq.~\eqref{eq:window_logits}
        \STATE $\{\mathcal{I}_{\mathrm{sel},h}^{(t+1)}\}_h \leftarrow \operatorname{Selector}(\{\ell_{\tau}\}, \mathcal{J}^{(t)}, K)$ using Alg.~\ref{alg:selector}
      \ENDIF
      \STATE Generate token $x_t$
      \STATE Update $\mathcal{I}_{\mathrm{recent}}^{(t+1)}$ and compute $g_{t+1}$ using Eq.~\eqref{eq:trigger} together with $T_{\max}$
      \STATE $t \leftarrow t+1$
    \ENDWHILE
  \end{algorithmic}
\end{algorithm}

\begin{algorithm}[t]
  \caption{{Selector}: closed-form fusion and log-score discretization (per layer)}
  \label{alg:selector}
  \begin{algorithmic}[1]
    \REQUIRE Window logits $\{\ell_{\tau}(j)\}_{\tau=1..W,\ j\in\mathcal{J}}$; cached key norms $\|k_h(j)\|_2$ and positions $u(j)$ for $j\in\mathcal{J}$; budget $K$; hyperparameters from Sec.~\ref{sec:selector}--\ref{sec:refine_topk}
    \ENSURE Selected indices $\mathcal{I}_{\mathrm{sel},h}\subseteq \mathcal{J}$ for each head $h$

    \STATE \textbf{Implementation note:} operations over heads $h$ and positions $j$ are independent and can be implemented as parallel tensor operations

    \STATE \textbf{(A) Continuous importance computation}
    \FOR{each head $h$}
      \STATE Compute $p_{\tau}(\cdot)$ on $\mathcal{J}$ using Eq.~\eqref{eq:selector_pt}, and aggregate $\mu(\cdot)$ using Eq.~\eqref{eq:selector_mu}
      \STATE Form the evidence distribution $f_h(\cdot)$ and prior distribution $r_h(\cdot)$ using Eq.~\eqref{eq:selector_f} and Eq.~\eqref{eq:selector_r}
      \STATE Choose $\lambda^*$ via Eq.~\eqref{eq:selector_lambda_closed}; then compute $s_h(\cdot)$ by Eq.~\eqref{eq:selector_closed} and initialize $z_h(\cdot)$ by Eq.~\eqref{eq:z_base}
    \ENDFOR

    \STATE \textbf{(B) Log-score refinement and discretization}
    \FOR{each head $h$}
      \STATE Apply Soft-NMS to $z_h(\cdot)$ using Eqs.~\eqref{eq:nms_pool}--\eqref{eq:nms}
    \ENDFOR
    \FOR{each position $j\in\mathcal{J}$}
      \STATE Apply cross-head exclusivity at position $j$ using Eqs.~\eqref{eq:head_softmax}--\eqref{eq:head_mutex}
    \ENDFOR
    \FOR{each head $h$}
      \STATE $\mathcal{I}_{\mathrm{sel},h}\leftarrow \mathrm{TopK}_{j\in\mathcal{J}}(z_h^{\mathrm{adj}}(j),K)$ using Eq.~\eqref{eq:topk_out}
    \ENDFOR
  \end{algorithmic}
\end{algorithm}

\paragraph{Algorithm discussion.}
Algorithm~\ref{alg:sfi_loop} implements the state machine in Sec.~\ref{sec:framework}: fast steps attend to the managed sparse state, while slow steps perform dense attention, record window logits, and invoke the Selector.
Algorithm~\ref{alg:selector} matches Sec.~\ref{sec:selector}--\ref{sec:refine_topk}: it builds evidence and prior on the shared support $\mathcal{J}$, fuses them via the exact reverse-KL closed form to obtain $s_h$, and then applies the discretization pipeline in log-score space (Soft-NMS, cross-head exclusivity, Top-$K$).
The refinements are introduced to increase within-head positional coverage and reduce cross-head redundancy at discretization time; they do not modify the probabilistic fusion objective in Sec.~\ref{sec:selector}.

\section{System Design and Kernel Optimization}
\label{sec:infra}

\begin{figure}[t]
  \centering
  \includegraphics[width=0.8\linewidth]{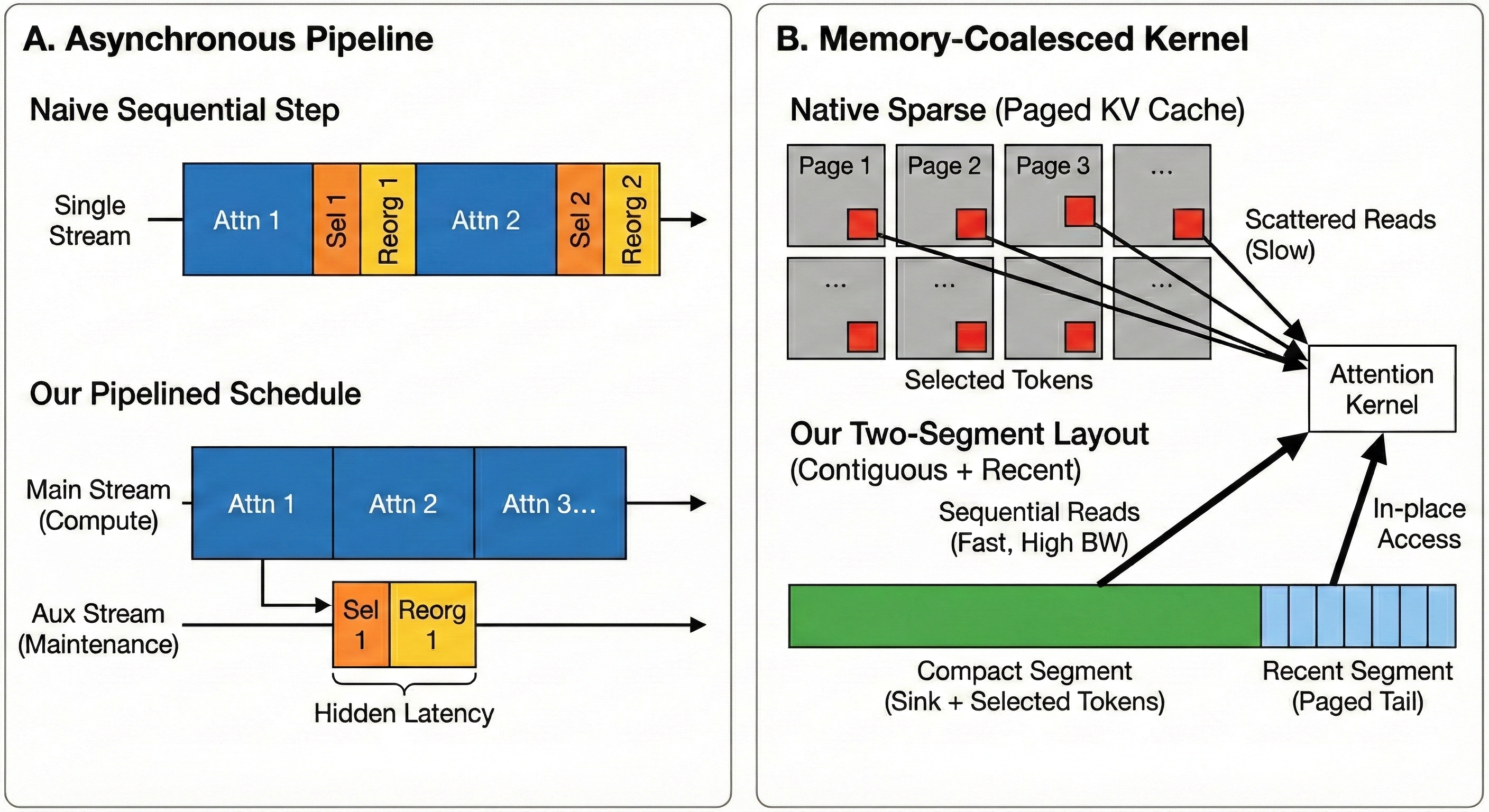}
 \caption{\textbf{System infrastructure for efficient Slow-Fast Inference.}
\textbf{(A) Asynchronous pipeline:} SFI overlaps the main attention computation with slow-step maintenance across two execution streams. While the Main Stream computes attention for layer $i+1$, the Aux Stream concurrently runs the Selector and cache reorganization for layer $i$, so that most maintenance overhead is hidden behind ongoing layer execution.
\textbf{(B) Memory-coalesced sparse kernel:} Native sparse attention suffers from scattered KV reads and poor bandwidth utilization. We therefore use a two-segment layout in which sink and selected tokens are packed into a contiguous compact buffer, enabling high-bandwidth sequential access over most of the sparse context, while recent tokens are read in place from paged KV.}
  \label{fig:infra_system}
\end{figure}

The Slow-Fast paradigm reduces attention FLOPs by replacing most full-history scans with attention over a compact sparse cache. Turning this algorithmic reduction into wall-clock speedup, however, requires addressing two system bottlenecks introduced by sparse reuse. First, a slow step does more than standard dense attention: it must also run the \emph{Selector} to refresh the sparse indices $\mathcal{I}_{\text{sel}}$, which can create a latency spike if handled synchronously. Second, sparse attention is not automatically efficient on GPUs. If fast steps read the selected KV pairs through irregular gathers from a paged cache, sparse attention can even become \emph{slower} than dense attention because of poor memory coalescing and bandwidth utilization. To address the latter issue, we reorganize the selected KV pairs after each slow step into a contiguous compact buffer for the subsequent fast steps. These two bottlenecks motivate the system design in Figure~\ref{fig:infra_system}: a layer-wise asynchronous schedule to hide slow-step maintenance latency, and memory-coalesced sparse kernels built on the reorganized compact buffer to make fast-step attention bandwidth-efficient.

\subsection{Hiding Slow-step Overhead via Layer-wise Asynchronous Execution}
\label{sec:infra_async}

A slow step traverses all $L$ transformer layers. For layer $i$, let $\mathrm{Attn}_i$ denote the full-attention computation, $\mathrm{Sel}_i$ the Selector computation, and $\mathrm{Reorg}_i$ the cache reorganization that materializes the sparse cache used by subsequent fast steps. Concretely, after $\mathrm{Sel}_i$ determines the refreshed selected indices for layer $i$, $\mathrm{Reorg}_i$ gathers the corresponding KV entries together with the sink segment into a contiguous compact buffer. This buffer is the layout consumed by the fast-path sparse kernel; recent tokens remain in paged KV and are handled separately, as illustrated in Figure~\ref{fig:infra_system}B.

With this role of reorganization made explicit, the dependency structure is simple. $\mathrm{Sel}_i$ depends on the evidence produced by $\mathrm{Attn}_i$, and $\mathrm{Reorg}_i$ in turn depends on the selected indices from $\mathrm{Sel}_i$. By contrast, the next-layer attention $\mathrm{Attn}_{i+1}$ does not depend on either $\mathrm{Sel}_i$ or $\mathrm{Reorg}_i$. Formally, the required edges are
\[
  \mathrm{Attn}_i \rightarrow \mathrm{Sel}_i \rightarrow \mathrm{Reorg}_i,
  \qquad
  \mathrm{Attn}_{i+1}\ \perp\ \{\mathrm{Sel}_i,\mathrm{Reorg}_i\}.
\]
This means that slow-step maintenance for layer $i$ can be overlapped with the main attention computation of later layers, rather than being placed directly on the critical path.

We implement this overlap with two CUDA streams (Figure~\ref{fig:infra_system}A). The primary stream runs $\mathrm{Attn}_{1},\ldots,\mathrm{Attn}_{L}$ sequentially, as in standard inference. As soon as $\mathrm{Attn}_i$ finishes, we launch $\mathrm{Sel}_i$ and $\mathrm{Reorg}_i$ on a secondary lower-priority stream, while the primary stream immediately proceeds to $\mathrm{Attn}_{i+1}$. In effect, selector and reorganization for one layer are pulled behind the full-attention computation of the following layers. Since full attention is typically the dominant per-layer cost in a slow step, this schedule hides most of the maintenance overhead under ongoing attention work and prevents slow steps from becoming pronounced latency spikes.

An efficient implementation should also avoid dynamic memory allocation and keep synchronization simple and predictable. We therefore use a fixed-size ring buffer with $R$ slots (typically $R=4$) to store the temporary tensors needed by the maintenance path, including the evidence consumed by $\mathrm{Sel}_i$ and the intermediate outputs produced by $\mathrm{Reorg}_i$. Layer $i$ writes to slot $(i \bmod R)$, and a CUDA event is recorded on the secondary stream immediately after the last maintenance operation that uses that slot. When the same slot is reused later by layer $i+R$, the secondary stream waits on the recorded event before reusing the slot. This ensures that a slot is recycled only after all earlier work on that slot has completed.

Finally, at the end of the slow step we insert a single completion barrier to ensure that all pending $\mathrm{Reorg}_i$ operations have finished before the next token starts Fast-Step decoding with the refreshed compact caches. In our experiments, this layer-wise overlap does not affect output quality, because the reorganized sparse cache is always completed before it is consumed by subsequent fast-path decoding.

\subsection{Memory-coalesced Kernel Design and Selector Maintenance}
\label{sec:infra_kernels}

Implementing the Slow-Fast paradigm requires specialized kernel support.
We note that standard highly optimized attention kernels (e.g., Flash-Attention~\citep{shah2024flashattention} or PyTorch SDPA~\citep{dao2024flashattention, xFormers2022}) are designed to output hidden states directly and typically do not expose the intermediate attention logits required by our Selector.
In our implementation, custom kernels are used both to accelerate fast step sparse attention over the compact cache and to expose the evidence logits required by the Selector efficiently.

\textbf{Coalesced reads with a two-segment KV layout.}
Sparse indexing alone does not guarantee speed. If fast steps fetch selected KV pairs through scattered reads from a paged KV cache~\citep{kwon2023efficient}, the resulting access pattern is bandwidth-inefficient and can offset the computational savings of sparse attention (Figure~\ref{fig:infra_system}B, top). We therefore use cache reorganization only for the long-range portion that will be reused across subsequent fast steps: during $\mathrm{Reorg}_i$, the sink and selected KV entries are packed into a contiguous compact buffer. fast-step attention can then read this long-range segment mostly through sequential, coalesced accesses, which is markedly more GPU-friendly than gather-based sparse reads (Figure~\ref{fig:infra_system}B, bottom).

Recent tokens are handled differently. Because the recent window changes at every step, explicitly repacking it into a separate contiguous buffer would introduce additional copies and window-maintenance overhead. Instead, our kernel reads the recent segment directly in place from the tail of the paged KV cache. The resulting Fast-Step kernel therefore operates on a two-segment input: a reorganized compact segment for sink and selected tokens, and an in-place paged segment for recent tokens. This design combines the bandwidth efficiency of coalesced reads on the reusable long-range cache with low maintenance overhead for the rapidly changing recent context.

\textbf{A GPU-native selector with a step-wise packed interface (Top-$K$ dominates).}
To keep slow-step maintenance lightweight, we implement the Selector as a fully GPU-resident pipeline. Probability aggregation, prior fusion, score construction, and the log-score refinements in Sec.~\ref{sec:refine_topk} are executed as vectorized device kernels, with all intermediate results kept on GPU. This avoids repeated host intervention, excessive kernel dispatch, and unnecessary materialization of large intermediate tensors. After fusion, the effective selector cost is dominated by the Top-$K$ primitive; the remaining computation is relatively light and highly parallel.

A second source of overhead in production runtimes is repeated per-layer metadata preparation and host-side control. To eliminate this, we construct a compact \emph{packed descriptor} once per decoding step, encoding the small control state required by both the Selector and sparse attention, such as per-request cache boundaries, block-table pointers, and layout offsets. Our custom kernels consume this descriptor directly from device memory. As a result, per-layer execution reduces to reading the packed descriptor and launching the corresponding kernels, avoiding CPU-side bookkeeping, CPU--GPU transfers, and extra synchronization on the hot path. This step-wise packing amortizes control overhead across layers and keeps the steady-state decode path largely kernel-only.

Overall, the system design in this section addresses both major sources of overhead introduced by Slow-Fast execution. Layer-wise asynchronous overlap prevents selector and cache reorganization from turning slow steps into latency spikes; the two-segment KV layout makes Fast-Step sparse attention bandwidth-efficient; and the GPU-native selector together with step-wise packed descriptors minimizes control and launch overhead. Taken together, these optimizations ensure that the algorithmic savings of SFI are realized in practice as consistent end-to-end decoding speedups.

\section{Experiments}

We first evaluate end-to-end throughput, then study task quality on long-context and long-CoT benchmarks, and finally analyze the contribution of the Selector through ablations.

\subsection{Experiment Setup}

\noindent\textbf{Models and evaluation scope.}
We evaluate task quality on three representative Qwen3~\citep{yang2025qwen3} checkpoints: Qwen3-4B, Qwen3-30B-A3B, and Qwen3-235B-A22B, covering small, medium, and large model scales. To study efficiency trends at smaller scales, we additionally include Qwen3-0.6B in the throughput evaluation. For each model, we compare the original full-KV baseline (full attention) with our proposed SFI. For head-to-head baseline comparisons, we focus on representative \textbf{training-free} KV-cache compression methods~\citep{li2024snapkv,zhang2023h2o,xiao2024efficient,cai2024pyramidkv,zhou2024dynamickv} with publicly available implementations.
In the tables below, we denote the full-KV baseline as \textbf{Slow}, since it corresponds to always using dense full attention at every decoding step.

\noindent\textbf{Benchmarks.}
We evaluate SFI under two long-sequence inference regimes. First, for \textit{long-context} settings, where the model must process large prefills and reason over extended input contexts, we use LongBench-V1~\citep{bai2024longbenchV1} and LongBench-V2~\citep{bai2025longbenchV2}. Second, for \textit{long-CoT} settings, where the model may generate long reasoning traces before producing the final answer, we use GPQA-Diamond~\citep{rein2024gpqa} and MMLU~\citep{hendrycks2024mmlu} with Qwen3 Thinking checkpoints. Throughout the section, ``long-CoT'' refers to this inference regime rather than an intrinsic property of the benchmark itself.

\noindent\textbf{Runtime environment and implementation.}
For Qwen3-4B and Qwen3-30B-A3B, both the full-KV baseline and SFI are evaluated on a single NVIDIA B200-192G GPU. For Qwen3-235B-A22B, both modes are evaluated on 8$\times$ NVIDIA B200-192G GPUs. All experiments are conducted with PyTorch 2.8.0 + cu128, vLLM 0.10.0, and Python 3.12. Since SFI is implemented on top of the Triton backend~\citep{tillet2019triton} in vLLM, we use the same Triton backend for the full-KV baseline to ensure a fair runtime comparison.

We report our main throughput results under vLLM because it is a high-performance inference runtime that more closely reflects realistic deployment settings. In such a runtime, measured speedups are less likely to come from incidental implementation differences, such as avoidable tensor copies or less optimized execution paths, and are therefore more indicative of the benefit of SFI itself.

\noindent\textbf{Baseline comparison protocol.}
For comparisons with other training-free KV-cache compression methods, we use their official implementations. As these methods do not provide vLLM-based versions, we evaluate them under NVIDIA’s \texttt{kvpress}~\citep{devoto2025expected} framework built on Hugging Face Transformers. To assess cross-framework consistency, we additionally verify on Qwen3-4B that both the full-KV baseline and SFI produce aligned quality results across vLLM and Transformers. This gives us confidence that the quality conclusions reported for SFI are not artifacts of a particular runtime. Accordingly, comparisons with other KV-cache compression methods are intended as \textit{task-quality} comparisons rather than wall-clock throughput comparisons, whereas all reported throughput numbers are measured only between SFI and the full-KV baseline under the same vLLM backend. For all KV-cache compression baselines evaluated under \texttt{kvpress}, we use a unified compression ratio of 0.5; by contrast, the average retained ratio of SFI is notably lower (approximately 15\%).

\noindent\textbf{Default SFI configuration.}
Unless otherwise specified, we use $|\mathcal{I}_{\mathrm{sink}}|=4$ sink tokens, a recent window of 256 tokens, and a selected-token budget of $K=2048$ per KV head. Slow steps are triggered by a tokenizer-level boundary set including \{\texttt{.}, \texttt{?}, \texttt{!}, \texttt{;}, \texttt{\textbackslash n}\}, together with a refresh budget of $T_{\text{max}}=64$. For decode we use $W=1$, while for prefill we use a tail window of $W=16$. The default refinement hyperparameters are $\lambda_{\text{clip}}=0.02$, $\alpha_{\text{soft}}=0.5$, and $\alpha_{\text{cross}}=0.35$. These defaults are used in all main experiments unless explicitly noted otherwise.

\begin{figure}[t!]
\centering
\includegraphics[width=0.9\linewidth]{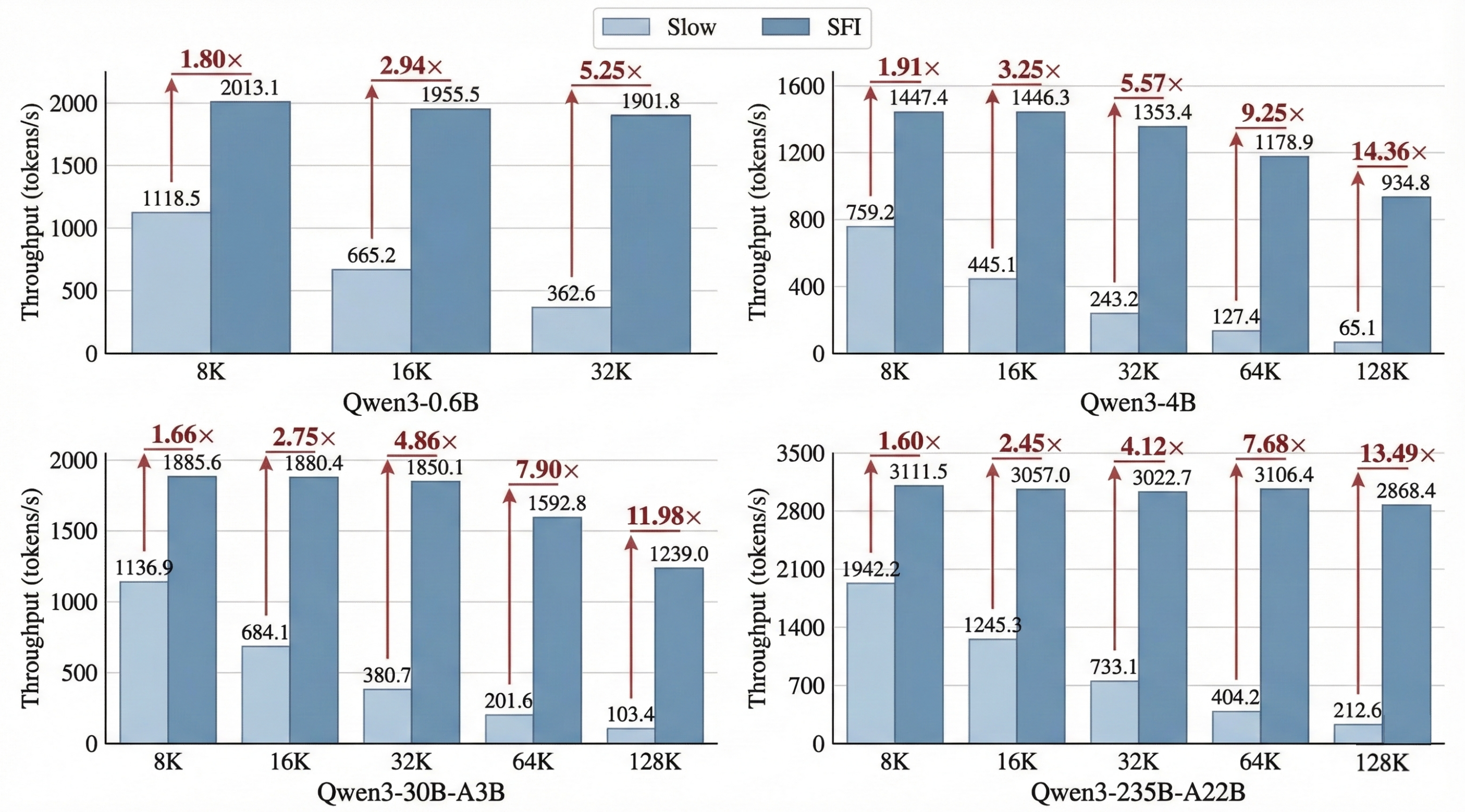}
\caption{\textbf{End-to-end decoding throughput across model scales and context lengths.}
Throughput (tok/s) is measured for the full-KV baseline (full kv cache) and SFI with up to 2048 generated tokens per request. SFI consistently improves decoding throughput, and the advantage grows with context length.}
\label{fig:speed_comparison}
\end{figure}

\subsection{Efficiency Analysis}

We evaluate decoding efficiency on four Qwen3 scales: Qwen3-0.6B, Qwen3-4B-Instruct-2507, Qwen3-30B-A3B, and Qwen3-235B-A22B. For the 0.6B, 4B, and 30B models, experiments are run on a single NVIDIA B200-192G GPU; for the 235B model, we use 8$\times$ NVIDIA B200-192G GPUs. All measurements are performed in bfloat16 under the same vLLM runtime~\citep{kwon2023efficient}. 
The full-KV baseline uses vLLM's standard dense-attention path, while SFI is implemented within the same vLLM inference stack and adds only the method-specific sparse kernels and maintenance logic required by Slow-Fast execution. This keeps the underlying implementation consistent across methods and makes the measured speedups more directly attributable to the decoding strategy itself rather than to differences in the  system.

For each setting, we generate up to 2048 new tokens per request with GPU memory utilization set to 0.8. We use batch sizes of 16 for the 0.6B and 4B models, 32 for the 30B-A3B model, and 128 for the 235B-A22B model, reflecting realistic serving settings where larger models are usually deployed with higher concurrency in order to better utilize GPU compute and amortize per-step runtime overhead. Context lengths range from 8K to 128K tokens, except for Qwen3-0.6B, which is limited to 32K by its positional-embedding ceiling (\texttt{max\_position\_embeddings}=40960). For each configuration, we perform one warm-up run followed by three measured runs and report the mean decoding throughput in tokens per second. The prefix cache is cleared before every run to avoid caching effects.

Figure~\ref{fig:speed_comparison} reports end-to-end decoding throughput. Across all model scales, SFI consistently outperforms the full-KV baseline, and the advantage becomes larger as the context grows. For Qwen3-4B, the speedup increases from $1.91\times$ at 8K to $14.36\times$ at 128K. Similar scaling trends are observed for Qwen3-30B-A3B ($1.66\times \!\to\! 11.98\times$) and Qwen3-235B-A22B ($1.60\times \!\to\! 13.49\times$). Even the smallest Qwen3-0.6B model reaches a $5.25\times$ throughput gain at 32K context. This trend aligns with the design of SFI: as context length grows, the cost of repeated full-history attention becomes increasingly dominant during decoding, while SFI reduces this burden by reusing sparse support across multiple steps.

Another notable pattern is that SFI degrades much more slowly with context length in absolute throughput. For example, Qwen3-4B sustains about 1400 tok/s from 8K to 32K and still reaches 935 tok/s at 128K, whereas the full-KV baseline drops from 759 tok/s to 65 tok/s over the same range. This shows that SFI makes decoding throughput markedly less sensitive to total context length by avoiding repeated attention over the full KV cache at most steps. The effect is also pronounced in the MoE models, where sparse expert activation complements SFI's sparse attention and leads to particularly high absolute throughput (e.g., 3111 tok/s for Qwen3-235B-A22B at 8K).

To isolate the effect of the sparse-attention implementation itself, Table~\ref{tab:kernel_speedup} reports kernel-level speedup at a fixed KV length of 16K under different retention ratios. These numbers are not end-to-end decoding throughput; they measure only the latency of a single sparse-attention kernel relative to its dense-attention counterpart. As the retained ratio decreases, the sparse kernel becomes substantially faster, reaching $10.67\times$ speedup at 1.6\% retention and approaching $1\times$ as the ratio approaches the dense case. This result shows not only that sparse attention can reduce computation effectively, but also that our memory-coalesced kernel with a two-segment KV layout turns that sparsity into real kernel-level acceleration on GPU. In particular, by storing the reusable long-range cache in a contiguous compact segment while reading the recent segment in place, the kernel avoids the bandwidth collapse that would arise from naive scattered gathers over paged KV. The remaining gap between kernel-level speedup and end-to-end throughput reflects additional costs from scheduling, Selector execution, cache maintenance, and other non-attention components. At the same time, these efficiency gains matter most precisely in the long-context regime where decode-time cost becomes most burdensome, which is increasingly relevant for modern reasoning and agentic workloads.

\begin{table}[t]
\centering
\caption{\textbf{Kernel-level sparse-attention speedup at KV length 16K.}
This measurement isolates the attention kernel itself rather than end-to-end decoding throughput. Speedup is computed as dense-attention latency divided by sparse-attention latency. Batch size = 16, bf16.}
\label{tab:kernel_speedup}
\renewcommand{\arraystretch}{1.2}
\setlength{\tabcolsep}{6pt}
\small
\begin{tabular}{c c c c c c c c c c}
\toprule
\textbf{Retention Ratio (\%)} & 1.6 & 6.3 & 12.5 & 25.0 & 37.5 & 50.0 & 75.0 & 98.4 & 100 \\
\midrule
\textbf{Speedup} & \textcolor{speedupgreen}{10.67$\times$} & \textcolor{speedupgreen}{9.56$\times$} & \textcolor{speedupgreen}{7.15$\times$} & \textcolor{speedupgreen}{3.96$\times$} & \textcolor{speedupgreen}{2.75$\times$} & \textcolor{speedupgreen}{2.10$\times$} & \textcolor{speedupgreen}{1.43$\times$} & 1.10$\times$ & 1.00$\times$ \\
\bottomrule
\end{tabular}
\end{table}

\begin{table*}[!t]
\centering
\caption{\textbf{LongBench-V1 results across Qwen3 scales.} SFI matches or improves full-KV decoding (Slow) on most subsets, with the clearest average gains at 4B and 30B-A3B. Bold marks the better result within each backbone; \textbf{\textcolor{deltag}{green}} values with $\uparrow$ denote gains over Slow.}
\label{tab:longbenchv1_results}
\footnotesize
\setlength{\tabcolsep}{8pt}
\renewcommand{\arraystretch}{1.15}
\begin{tabular}{@{} l l c >{\columncolor{sfiblue}}c !{\hspace{6pt}\vrule width 0.6pt\hspace{6pt}} c >{\columncolor{sfiblue}}c !{\hspace{6pt}\vrule width 0.6pt\hspace{6pt}} c >{\columncolor{sfiblue}}c}
\Xhline{1.2pt}
\rule{0pt}{3.0ex}
& & \multicolumn{2}{c}{\textbf{Qwen3-4B}} & \multicolumn{2}{c}{\textbf{Qwen3-30B-A3B}} & \multicolumn{2}{c}{\textbf{Qwen3-235B-A22B}} \\
\cmidrule(lr){3-4} \cmidrule(lr){5-6} \cmidrule(lr){7-8}
\rule{0pt}{2.4ex}
\textbf{Category} & \textbf{Task} & Slow & \textbf{SFI (Ours)} & Slow & \textbf{SFI (Ours)} & Slow & \textbf{SFI (Ours)} \\
\midrule
%
\multirow{3}{*}{\textit{Single-Doc QA}}
& Qasper          & 40.60 & \textbf{44.20} \gain{3.6}  & 38.96 & \textbf{42.14} \gain{3.2}  & \textbf{45.77} & 44.67 \\
& MultiFieldQA-en & 46.56 & \textbf{49.31} \gain{2.8}  & 50.32 & \textbf{53.22} \gain{2.9}  & \textbf{51.21} & 50.92 \\
& MultiFieldQA-zh & 61.21 & \textbf{63.81} \gain{2.6}  & 63.42 & \textbf{66.00} \gain{2.6}  & \textbf{67.08} & 66.37 \\
\cmidrule(lr){1-8}
%
\multirow{4}{*}{\textit{Multi-Doc QA}}
& HotpotQA        & 55.34 & \textbf{59.00} \gain{3.7}  & 61.68 & \textbf{63.37} \gain{1.7}  & 67.13 & \textbf{67.65} \gain{0.5} \\
& 2WikiMQA        & 42.02 & \textbf{44.50} \gain{2.5}  & 54.68 & \textbf{55.98} \gain{1.3}  & 64.14 & \textbf{65.31} \gain{1.2} \\
& MuSiQue         & 24.79 & \textbf{25.76} \gain{1.0}  & \textbf{32.22} & 31.95              & 42.86 & \textbf{43.44} \gain{0.6} \\
& DuReader        & 21.85 & \textbf{23.77} \gain{1.9}  & 21.40 & \textbf{23.44} \gain{2.0}  & \textbf{25.38} & 24.58 \\
\cmidrule(lr){1-8}
%
\multirow{3}{*}{{\textit{Summarization}}}
& GovReport       & 27.87 & \textbf{29.72} \gain{1.9}  & 29.40 & \textbf{30.18} \gain{0.8}  & \textbf{31.68} & 31.28 \\
& QMSum           & 22.11 & \textbf{22.21}  & 21.66 & \textbf{21.92}  & \textbf{22.81} & 22.72 \\
& MultiNews       & \textbf{24.06} & 24.04              & \textbf{23.52} & 23.46              & \textbf{23.46} & 23.33 \\
\cmidrule(lr){1-8}
%
\multirow{4}{*}{\textit{Few-Shot Learning}}
& TREC            & 73.00 & \textbf{75.00} \gain{2.0}  & 77.50 & \textbf{78.50} \gain{1.0}  & \textbf{77.50} & \textbf{77.50} \\
& TriviaQA        & 85.29 & \textbf{85.62}  & \textbf{91.56} & 91.06              & 91.86 & \textbf{92.10} \\
& SAMSum          & 39.12 & \textbf{39.99} \gain{0.9}  & 39.12 & \textbf{39.72} \gain{0.6}  & 41.09 & \textbf{41.30} \\
& LSHT            & 30.25 & \textbf{37.75} \gain{7.5}  & 42.50 & \textbf{47.25} \gain{4.8}  & 51.00 & \textbf{52.00} \gain{1.0} \\
\cmidrule(lr){1-8}
%
\multirow{3}{*}{\textit{Synthetic \& Code}}
& PassageRet-en   & \textbf{100.0} & \textbf{100.0}     & \textbf{100.0} & \textbf{100.0}     & \textbf{100.0} & \textbf{100.0} \\
& LCC             & \textbf{4.48} & 4.32                 & 24.82 & \textbf{25.13}  & \textbf{61.32} & 61.10 \\
& RepoBench-P     & 5.17 & \textbf{5.28}      & 24.76 & \textbf{24.92}  & 62.56 & \textbf{63.18} \gain{0.6} \\
\Xhline{0.8pt}
\rule{0pt}{2.6ex}
\textbf{Average} & & 41.40 & \textbf{43.19} \gain{1.8}  & 46.91 & \textbf{48.13} \gain{1.2}  & 54.52 & \textbf{54.56} \\
\Xhline{1.2pt}
\end{tabular}
\end{table*}

\begin{table*}[t]
\centering
\caption{\textbf{LongBench-V2 results across Qwen3 scales.} SFI remains stable across model sizes and shows the strongest gain on the longest-context subset of Qwen3-235B-A22B. Bold marks the better result within each backbone; \textbf{\textcolor{deltag}{green}} values with $\uparrow$ denote gains over Slow.}
\label{tab:longbenchv2_results}
\small
\setlength{\tabcolsep}{10pt}
\renewcommand{\arraystretch}{1.15}
\begin{tabular}{@{} l l cccccc }
\Xhline{1.2pt}
\rule{0pt}{2.6ex}
\textbf{Backbone} & \textbf{Method} & \textbf{Overall} & \textbf{Easy} & \textbf{Hard} & \textbf{Short} & \textbf{Medium} & \textbf{Long} \\
\midrule
\multirow{2}{*}{\textbf{Qwen3-4B}}
& Slow & 34.2 & 37.5 & 32.2 & \textbf{35.0} & 33.5 & \textbf{34.3} \\
& \cellcolor{sfiblue}\textbf{SFI (Ours)} & \cellcolor{sfiblue}\textbf{34.8} \gain{0.6} & \cellcolor{sfiblue}\textbf{38.5} \gain{1.0} & \cellcolor{sfiblue}\textbf{32.5} & \cellcolor{sfiblue}\textbf{35.0} & \cellcolor{sfiblue}\textbf{34.9} \gain{1.4} & \cellcolor{sfiblue}\textbf{34.3} \\
\cmidrule(lr){1-8}
\multirow{2}{*}{\textbf{Qwen3-30B-A3B}}
& Slow & \textbf{35.1} & \textbf{36.0} & \textbf{34.6} & \textbf{38.5} & \textbf{32.8} & \textbf{28.6} \\
& \cellcolor{sfiblue}\textbf{SFI (Ours)} & \cellcolor{sfiblue}\textbf{35.1} & \cellcolor{sfiblue}\textbf{36.0} & \cellcolor{sfiblue}\textbf{34.6} & \cellcolor{sfiblue}\textbf{38.5} & \cellcolor{sfiblue}\textbf{32.8} & \cellcolor{sfiblue}\textbf{28.6} \\
\cmidrule(lr){1-8}
\multirow{2}{*}{\textbf{Qwen3-235B-A22B}}
& Slow & \textbf{46.0} & \textbf{50.0} & \textbf{43.7} & \textbf{48.0} & \textbf{44.1} & 47.6 \\
& \cellcolor{sfiblue}\textbf{SFI (Ours)} & \cellcolor{sfiblue}\textbf{46.0} & \cellcolor{sfiblue}\textbf{50.0} & \cellcolor{sfiblue}\textbf{43.7} & \cellcolor{sfiblue}47.5 & \cellcolor{sfiblue}\textbf{44.1} & \cellcolor{sfiblue}\textbf{52.4} \gain{4.8} \\
\Xhline{1.2pt}
\end{tabular}
\vspace{-0.5em}
\end{table*}

\begin{table*}[!t]
\centering
\caption{\textbf{LongBench-V2 comparison with representative training-free KV-cache compression baselines on Qwen3-4B-Instruct-2507.} All baselines use 50\% compression, whereas SFI retains only 15--20\% of tokens on average and still achieves the best overall score. Bold indicates the best result and underline indicates the second-best result among compression methods in each column.}
\label{tab:longbench-v2}
\setlength{\tabcolsep}{9pt}
\renewcommand{\arraystretch}{1.15}
\begin{tabular}{ l cc ccc c }
\Xhline{1.2pt}
\rule{0pt}{2.8ex}
& \multicolumn{2}{c}{\textbf{Difficulty}} & \multicolumn{3}{c}{\textbf{Context Length}} & \\
\cmidrule(lr){2-3} \cmidrule(lr){4-6}
\rule{0pt}{2.2ex}
\textbf{Method} & Easy & Hard & Short & Medium & Long & \textbf{Avg.} \\
\midrule
Slow            & 37.50 & 32.20 & 35.00 & 33.50 & 34.30 & 34.20 \\
\cmidrule(lr){1-7}
StreamingLLM~\citep{xiao2024efficient}  & \textbf{39.58} & 26.69 & \textbf{36.67} & 28.84 & 28.70 & 31.61 \\
TOVA~\citep{oren2024transformers}       & \underline{38.54} & 27.65 & \underline{36.11} & 31.63 & 25.00 & 31.81 \\
Think~\citep{xuthink}                  & 36.98 & 27.65 & 31.67 & 31.63 & \underline{29.63} & 31.21 \\
SnapKV~\citep{li2024snapkv}            & 36.98 & \underline{29.26} & \underline{36.11} & \underline{32.56} & 25.00 & 32.21 \\
ChunkKV~\citep{liu2025chunkkv}         & 36.98 & 28.94 & 33.89 & \underline{32.56} & 27.78 & 32.01 \\
LagKV~\citep{liang2025lagkv}           & 33.85 & 27.33 & 33.89 & 29.77 & 23.15 & 29.82 \\
KNorm~\citep{devoto2024simple}         & 28.13 & 24.12 & 26.67 & 25.58 & 24.07 & 25.65 \\
PyramidKV~\citep{cai2024pyramidkv}     & 38.02 & \underline{29.26} & \textbf{36.67} & 32.09 & 26.85 & \underline{32.60} \\
\cmidrule(lr){1-7}
\rowcolor{sfiblue}
\textbf{SFI (Ours)} & \underline{38.50} & \textbf{32.50} & 35.00 & \textbf{34.90} & \textbf{34.30} & \textbf{34.80} \\
\Xhline{1.2pt}
\end{tabular}
\vspace{-1em}
\end{table*}

\begin{table}[t]
\centering
\caption{\textbf{Long-CoT reasoning results on GPQA and MMLU.} SFI matches the full-KV baseline at medium and large scales, with only minor variation at 4B. Bold marks the better result within each model.}
\label{tab:long_cot_results}
\begin{tabular}{l|cc|cc}
\toprule
\textbf{Model} & \multicolumn{2}{c|}{\textbf{GPQA}} & \multicolumn{2}{c}{\textbf{MMLU}} \\
\cmidrule(lr){2-3} \cmidrule(lr){4-5}
& Slow & \cellcolor{sfiblue} SFI (Ours) & Slow & \cellcolor{sfiblue} SFI (Ours) \\
\midrule
Qwen3-4B-Thinking-2507 & \textbf{64.14} & \cellcolor{sfiblue} 63.70 & \textbf{63.00} & \cellcolor{sfiblue} \textbf{63.00} \\
Qwen3-30B-A3B-Thinking-2507 & 69.70 & \cellcolor{sfiblue} \textbf{71.21} & \textbf{70.90} & \cellcolor{sfiblue} 70.60 \\
Qwen3-235B-A22B-Thinking-2507 & \textbf{80.80} & \cellcolor{sfiblue} \textbf{80.80} & 90.09 & \cellcolor{sfiblue} \textbf{90.30} \\
\bottomrule
\end{tabular}
\end{table}

\subsection{Effectiveness Analysis}
Unless otherwise stated, the results in this subsection use the default SFI configuration described above, except that we increase the per-head selected-token budget from $K=2048$ to $K=3836$ for the main effectiveness evaluations.

We focus on long-sequence regimes, where reusing a sparse support across multiple decoding steps has the greatest opportunity to amortize the cost of dense attention. We therefore study two settings: long-context tasks with very long prefills, and long-CoT tasks with long generated reasoning traces. We first compare SFI against the Slow full-KV baseline, and then against representative training-free KV-cache compression methods.

\noindent\textbf{Long Context.}
Across LongBench-V1 and LongBench-V2, the overall pattern is consistent: SFI matches the full-KV baseline across scales, and in several cases improves performance, especially on harder or longer inputs. We evaluate three Qwen3 scales on LongBench-V1~\citep{bai2024longbenchV1} and LongBench-V2~\citep{bai2025longbenchV2}. The maximum context length is set to 252,144 tokens (approximately 256K). For multiple-choice tasks in LongBench-V2, we follow the official setting and limit the output length to 128 tokens.

On LongBench-V1 (Table~\ref{tab:longbenchv1_results}), which covers five major categories and 17 subsets, SFI is consistently competitive with the full-KV baseline and improves the average score at small and medium scales. For Qwen3-4B, the overall average increases from 41.40 to 43.19 (+1.8). For Qwen3-30B-A3B, it improves from 46.91 to 48.13 (+1.2). For Qwen3-235B-A22B, the overall score is essentially unchanged (54.52 to 54.56), while several multi-document reasoning subsets still improve. These results indicate that replacing full-history attention at every step with sparse support reuse does not compromise long-context quality, and can even help smaller and mid-sized models.

On LongBench-V2 (Table~\ref{tab:longbenchv2_results}), the same trend remains clear. For Qwen3-4B, the overall score improves from 34.2 to 34.8 (+0.6), with gains on the Easy and Medium subsets. Qwen3-30B-A3B is unchanged at the reported precision across all subsets. For Qwen3-235B-A22B, SFI preserves the overall score exactly (46.0 to 46.0) while improving the Long subset from 47.6 to 52.4 (+4.8). This behavior may stem from the fact that under very long contexts, the Selector keeps the most informative support tokens while filtering out many low-value or distracting ones, so the effective context seen by the model can become cleaner than using the entire KV cache indiscriminately.

We further compare SFI with representative training-free KV-cache compression methods on Qwen3-4B-Instruct-2507 (Table~\ref{tab:longbench-v2}). We evaluate these baselines under NVIDIA's \texttt{kvpress}~\citep{devoto2025expected}. In this comparison, all competing methods use a unified compression ratio of 0.5, whereas SFI retains only about 15--20\% of tokens on average. Despite operating with a much tighter token budget, SFI achieves the best overall score (34.80), is the only method that surpasses the Slow full-KV baseline in average accuracy (34.20), and delivers the strongest results on the Hard and Medium subsets. This gap suggests that selector quality matters at least as much as retention ratio: keeping fewer but better-targeted tokens can be more effective than retaining a much larger but less selective cache.

Taken together, these long-context results support a simple conclusion: SFI substantially reduces the active KV state and the frequency of dense attention while preserving accuracy close to the full-KV baseline. When gains do appear, we view them as a useful byproduct of better support selection, especially in long and noisy contexts, rather than the primary claim of the paper.

\noindent\textbf{Long CoT.}
For long-CoT reasoning, the main takeaway is robustness. We evaluate GPQA~\citep{rein2024gpqa} and MMLU~\citep{hendrycks2024mmlu} using Qwen3 Thinking checkpoints, with a maximum output length of 32K tokens and a maximum context length of 256K.

As shown in Table~\ref{tab:long_cot_results}, SFI remains stable for medium and large thinking models. For Qwen3-4B, GPQA changes from 64.14 to 63.70, while MMLU remains unchanged at 63.00; although the GPQA difference is visible in percentage terms, it effectively corresponds to only about one additional mistake on this relatively small evaluation set. For Qwen3-30B-A3B, SFI improves GPQA from 69.70 to 71.21 while remaining close on MMLU (70.90 vs.\ 70.60). For Qwen3-235B-A22B, SFI matches GPQA exactly (80.80 vs.\ 80.80) and slightly improves MMLU (90.30 vs.\ 90.09). Overall, these results suggest that sparse support reuse remains reliable even when the generated reasoning trace is long, while keeping performance close to the full-KV baseline once model scale is moderate.

\begin{table}[t]
\centering
\caption{\textbf{Effect of the clip threshold $\lambda_\text{clip}$ on LongBench-V1 category averages.} Performance is stable in the small-$\lambda_\text{clip}$ regime and degrades as $\lambda_\text{clip}$ becomes larger, indicating that mild prior clipping is sufficient while stronger prior injection becomes over-regularized. \textbf{Bold} denotes the best result per column; the shaded row marks the chosen configuration.}
\label{tab:ablation_clip}
\begin{threeparttable}
\begin{tabular}{c|ccccccc}
\toprule
$\lambda_\text{clip}$ &
S-QA & M-QA & Summ. & Few-shot & Synthetic & Code & \textbf{Avg.} \\
\midrule
0.005 & 43.05 & \textbf{50.32} & 21.64 & \textbf{64.24} & 72.33 & 24.74 & 46.05 \\
\rowcolor{gray!15}
0.02 & \textbf{43.14} & 50.29 & \textbf{21.73} & 64.21 & 72.40 & 24.83 & \textbf{46.10}  \\
0.05  & 43.02 & 50.27 & 21.60 & 64.21 & 72.33 & 24.81 & 46.04 \\
0.15  & 42.95 & 50.20 & 21.53 & 64.00 & 72.33 & 24.89 & 45.98 \\
0.25  & 42.60 & 49.95 & 21.66 & 63.85 & 72.17 & \textbf{24.90} & 45.86 \\
0.40  & 42.23 & 50.28 & 21.58 & 63.48 & \textbf{72.50} & 24.69 & 45.79 \\
\bottomrule
\end{tabular}
\end{threeparttable}
\end{table}

\begin{table}[t]
\centering
\caption{\textbf{Effect of score-refinement hyperparameters on LongBench-V1.} Moderate cross-head competition and moderate Soft-NMS produce the best overall result, supporting refinement that improves support diversity without suppressing useful tokens too aggressively. \textbf{Bold} denotes the best result per column; the shaded rows mark the chosen configuration.}
\label{tab:ablation_score_refinement}
\begin{tabular}{cc|ccccccc}
\toprule
$\alpha_{\text{cross}}$ & $\alpha_{\text{soft}}$ &
S-QA & M-QA & Summ. & Few-shot & Synthetic & Code & \textbf{Avg.} \\
\midrule
\multicolumn{9}{l}{\textit{(a)~Varying $\alpha_{\text{cross}}$ with $\alpha_{\text{soft}} = 0.0$ fixed}} \\
\midrule
0.00 & 0.0 & 43.21 & 50.02 & 21.80 & 64.39 & 72.35 & 25.22 & 46.52 \\
0.15 & 0.0 & 42.98 & 50.55 & 21.62 & \textbf{64.44} & 72.48 & 24.90 & 46.55 \\
0.25 & 0.0 & 43.44 & \textbf{50.63} & 21.66 & 64.26 & 72.33 & 24.82 & 46.61 \\
\rowcolor{gray!15}
0.35 & 0.0 & \textbf{43.45} & 50.29 & \textbf{21.84} & \textbf{64.44} & 72.33 & 25.06 & \textbf{46.70} \\
0.50 & 0.0 & 43.19 & 50.11 & 21.62 & 64.29 & 72.19 & 24.82 & 46.45 \\
0.65 & 0.0 & 43.04 & 50.42 & 21.56 & 64.17 & \textbf{72.67} & 25.01 & 46.51 \\
0.80 & 0.0 & 43.03 & 50.30 & 21.61 & 64.03 & 72.02 & 24.37 & 46.32 \\
0.95 & 0.0 & 43.24 & 50.34 & 21.68 & 64.41 & \textbf{72.67} & \textbf{25.07} & 46.64 \\
\midrule
\multicolumn{9}{l}{\textit{(b)~Varying $\alpha_{\text{soft}}$ with $\alpha_{\text{cross}} = 0.35$ fixed}} \\
\midrule
0.35 & 0.0 & \textbf{43.45} & 50.29 & \textbf{21.84} & 64.44 & 72.33 & 25.06 & 46.70 \\
\rowcolor{gray!15}
0.35 & 0.5 & 43.44 & \textbf{50.58} & 21.74 & \textbf{64.45} & \textbf{72.67} & 24.99 & \textbf{46.77} \\
0.35 & 1.0 & 43.43 & 50.25 & 21.82 & 64.42 & 72.31 & \textbf{25.08} & 46.65 \\
\bottomrule
\end{tabular}
\end{table}

\subsection{Ablation Study}
To isolate the contribution of different Selector components, we conduct ablations on Qwen3-30B-A3B-Instruct-2507 under long-context settings. Scores are averaged over the six task categories of LongBench-V1.

\noindent\textbf{Impact of $\lambda_{\text{clip}}$.}
Table~\ref{tab:ablation_clip} studies the effect of the clip threshold $\lambda_{\text{clip}}$.
The key pattern is that performance is stable once $\lambda_{\text{clip}}$ is sufficiently small, rather than improving monotonically as $\lambda_{\text{clip}}$ decreases. In particular, the low-$\lambda_{\text{clip}}$ regime already forms a plateau: $\lambda_{\text{clip}}=0.02$ achieves the best average score (46.10), while $\lambda_{\text{clip}}=0.05$ is nearly identical (46.04). As $\lambda_{\text{clip}}$ increases further, the average score gradually drops to 45.79 at 0.40. This trend suggests that a mild prior constraint is sufficient to stabilize the fusion between slow-step attention evidence and structural priors, whereas overly aggressive clipping injects too much prior bias and starts to distort the support preferred by dense attention.

\noindent\textbf{Impact of Score Refinement ($\alpha_{\text{cross}}$ and $\alpha_{\text{soft}}$).}
Table~\ref{tab:ablation_score_refinement} examines the two refinement terms in the Selector.
With $\alpha_{\text{soft}}=0$, moderate cross-head competition gives the best result: $\alpha_{\text{cross}}=0.35$ reaches the highest average score (46.70), while both weaker and stronger competition lead to small degradations. Fixing $\alpha_{\text{cross}}=0.35$, introducing moderate Soft-NMS further improves the average to 46.77 at $\alpha_{\text{soft}}=0.5$, whereas disabling it or strengthening it to 1.0 is slightly worse. Together, these results indicate that score refinement is most effective when it encourages diversity without over-suppressing high-confidence candidates: too little refinement leaves redundant support across heads, while too much refinement can remove tokens that remain genuinely useful.

These ablations clarify why SFI remains accurate even though fast steps attend only to a sparse support. The Selector must balance three competing objectives: fidelity to slow-step attention, controlled use of structural priors, and diversity of the retained support. $\lambda_{\text{clip}}$ governs how much prior information can influence the fused score; keeping it in a small stable regime prevents the prior from overwhelming dense-attention evidence. Meanwhile, $\alpha_{\text{cross}}$ and $\alpha_{\text{soft}}$ reduce redundant token allocation across heads and nearby positions, which improves coverage under a fixed token budget. This balance is especially important because the selected memory is reused across an entire fast segment. Once the support is calibrated well, reuse remains reliable for many subsequent steps; if it is poorly calibrated, the error would persist across the whole segment.

\section{Conclusion}
We presented SFI, a training-free decoding framework built on a simple observation: attention support often changes more slowly than tokens are generated. SFI exploits this structure through a slow-fast decoding schedule that alternates dense refresh at slow steps with sparse reuse at fast steps, using the Selector to convert slow-step evidence into reusable selected memory.
Across long-context and long-CoT settings, SFI preserves near-parity quality with the full-KV baseline while delivering substantial throughput gains, with the clearest benefits appearing as context length grows. These results suggest that the key opportunity is not to sparsify every decoding step in the same way, but to exploit the temporal stability of attention support across steps.
Because it applies directly to existing checkpoints without retraining, SFI offers a practical path to lowering inference cost for contemporary autoregressive reasoning models, especially in long-horizon and agentic workloads.

{\small
\bibliography{ref}
}

\end{document}